\documentclass[letterpaper]{article}
\usepackage[preprint]{aaai2027}
\usepackage[hyphens]{url}
\usepackage{graphicx}
\usepackage{natbib}
\usepackage{caption}
\usepackage{algorithm}
\usepackage{algorithmic}
\usepackage{newfloat}
\usepackage{listings}
\DeclareCaptionStyle{ruled}{labelfont=normalfont,labelsep=colon,strut=off}
\floatstyle{ruled}
\newfloat{listing}{tb}{lst}{}
\floatname{listing}{Listing}
\usepackage{booktabs}
\usepackage{amsmath}
\usepackage{multirow}
\usepackage{amssymb}
\title{ErodedFace: An Efficient and Robust Portrait Video Defense Framework \\
against unauthorized 3D Talking Face Generation}
\author{
Rui-qing Sun\textsuperscript{1,$\dagger$}\quad
Xingshan Yao\textsuperscript{1,$\dagger$}\quad
Tian Lan\textsuperscript{1}\\
Jia-Ling Shi\textsuperscript{1}\quad
Chen-Hao Cui\textsuperscript{1}\quad
Hui-Yang Zhao\textsuperscript{1}\\
Zhijing Wu\textsuperscript{1}\quad
Chen Yang\textsuperscript{1,*}\quad
Xian-Ling Mao\textsuperscript{1,*}
}
\affiliations{\textsuperscript{1}Beijing Institute of Technology}
\begin{document}
\maketitle
\begingroup
\renewcommand{\thefootnote}{\fnsymbol{footnote}}
\footnotetext[2]{These authors contributed equally.}
\footnotetext[1]{Corresponding authors.}
\endgroup
\begin{abstract}

Talking Face Generation (TFG) has achieved remarkable progress in synthesizing realistic speaker videos, and has evolved into two representative paradigms from the perspective of identity incorporation: image-based 2D TFG and video-based 3D TFG.
While TFG promotes efficient content creation, it also raises severe risks of unauthorized portrait forgery and privacy abuse, motivating the rapid development of proactive defense methods against unauthorized TFG.
Currently, the proactive defense methods mainly focus on unauthorized image-based 2D TFG. However, as far as we know, there is NO any proactive defense method for unauthorized video-based 3D TFG.
Intuitively, a portrait video can be decomposed into individual frames, to which existing defenses developed for unauthorized image-based 2D TFG are applied independently.
Experiments shows that this straightforward adaptation is both ineffective and computationally prohibitive: the vulnerable components exploited by existing defenses, such as image-reference modules and reverse diffusion processes, are absent from video-based 3D TFG pipelines, while per-frame optimization incurs substantial computational overhead.
To address this gap, we first analyzed the vulnerabilities of video-based 3D methods and then proposed an efficient proactive defense framework named ErodedFace.
Specifically, ErodedFace adopts a curriculum-inspired two-stage strategy that
targets the reference-video preprocessing phase in video-based 3D TFG: Stage I efficiently establishes scale-robust disruption through similarity-guided parameter inheritance and
adaptive multi-scale optimization, and Stage II further improves perceptual
fidelity and transformation robustness without weakening the established
defense.
Extensive experiments demonstrate that ErodedFace effectively protects portrait videos against unauthorized video-based 3D TFG, achieves a 10.0$\times$ speedup in defense generation compared with existing image-based 2D TFG defenses, and remains robust to common transformations and recent purification methods.

\end{abstract}

\section{Introduction}
\label{sec:intro}

Talking Face Generation (TFG) aims to synthesize realistic speaker videos of a target person conditioned on driving signals such as speech audio.
With the rapid progress of generative models, TFG has evolved into two
representative paradigms from the perspective of identity incorporation:
reference-image-conditioned generator, which directly consumes and
animates a static reference portrait during inference, and video-based
3D rendering ~\cite{Athar_2022_RigNeRF,Gu_2022_styleNerf,guo2021adnerf,Ye_2024_geneface,Peng_2024_synctalk,cho2024gaussiantalker,Li_2024_talkingaussian,Li_2025_instag}, which reconstructs a person-specific NeRF or 3DGS
representation from a reference portrait video and subsequently renders
audio-driven speaker videos from the learned representation.

While these techniques facilitate convenient content creation, they also expose personal portrait videos to severe risks of unauthorized portrait forgery and privacy abuse.
The increasing misuse risk of TFG has motivated the development of proactive protection techniques~\cite{liang2023AdvDM-,xue2023AdvDM+_SDS,salman2023PhotoGuard,liang2023Mist,gan2025silence}.
Existing defenses have been mainly designed for reference-image-conditioned generation, where the protected reference image is directly consumed by the generator during forgery.
In such pipelines, image-level perturbations can be injected into the reference image to interfere the final forgery process.

\begin{figure*}[!t]
    \centering
    \includegraphics[width=0.85\textwidth]{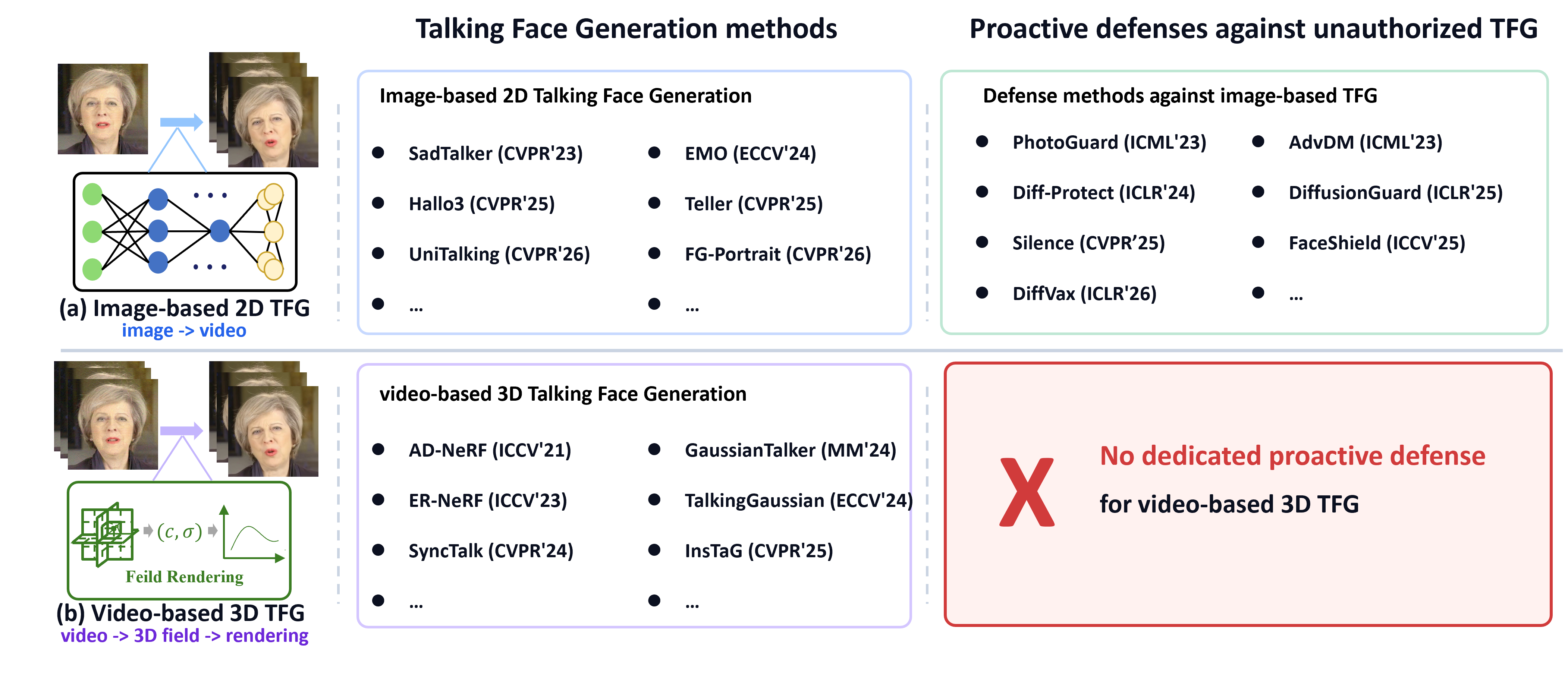}
    \caption{
    ErodedFace is, to the best of our knowledge, the first proactive defense designed for unauthorized video-based 3D TFG.
Existing proactive defenses mainly target image-based 2D TFG, where the protected reference image is directly consumed during generation.
In contrast, video-based 3D TFG follows a fundamentally different paradigm, as shown in (a) and (b).}
    \label{fig:motivation}
\end{figure*}

However, to the best of our knowledge, no proactive defense has been specifically designed for against video-based 3D rendering TFG, as illustrated in Fig.~\ref{fig:motivation}.
Directly applying existing reference-image-conditioned proactive defenses to portrait videos faces two major limitations. First, these defenses are architecturally incompatible with video-based 3D rendering TFG. They are designed to disrupt the direct conditioning path between a reference image and an image encoder, VAE, or diffusion model, whereas video-based 3D rendering TFG processes a reference video to reconstruct a person-specific NeRF or 3DGS representation before rendering. Consequently, their optimization objectives cannot effectively disrupt the personalized 3D representation reconstruction and rendering process. Second, their frame-wise application requires independently optimizing every video frame, causing the computational cost to increase linearly with the video length and making portrait video protection prohibitively time-consuming.

To bridge this protection gap, we propose \textbf{ErodedFace}, to the best of our knowledge, the first proactive defense framework specifically designed for against video-based 3D rendering TFG.
ErodedFace aims to degrade the rendering quality of unauthorized synthesis while preserving the perceptual fidelity of protected portrait videos.
Guided by these two objectives, we analyze the reference-video preprocessing phase and the video forgery pipeline of video-based 3D TFG, identifying intrinsic vulnerabilities that can be exploited for proactive protection.
Based on this analysis, ErodedFace introduces visually imperceptible adversarial perturbations into portrait videos and adopts a curriculum-learning-driven two-stage strategy to progressively optimize them.
By disrupting the reference-video preprocessing phase, ErodedFace prevents 3D TFG models from faithfully acquiring target-specific appearance and geometry cues, thereby compromising the downstream personalized 3D representation reconstruction and rendering process.

Specifically, ErodedFace adopts a curriculum-learning-driven two-stage defense strategy that progresses from shared-parameter multi-scale disruption to artifact-suppressed robust refinement.
The first stage leverages the temporal redundancy among adjacent video frames to share  multi-scale  perturbation parameters across the portrait video, which substantially accelerates adversarial optimization while inducing consistent disruption.
Despite its efficiency, the parameter-sharing mechanism may introduce redundant perturbation artifacts that degrade the perceptual fidelity of protected videos and weaken purification robustness.
Therefore, the second stage performs artifact-suppressed robust refinement to explicitly suppress redundant perturbation artifacts introduced by shared parameters, improving transformation and compression robustness while maintaining high perceptual fidelity.
Through this progressive disruption-and-refinement design, ErodedFace achieves an effective balance among defense effectiveness, perceptual fidelity, transformation and compression robustness, and optimization efficiency.

Extensive experiments demonstrate that ErodedFace provides effective and robust protection against unauthorized video-based 3D rendering TFG.
Compared with existing image-conditioned defense methods, ErodedFace more effectively degrades the rendering quality of unauthorized video-based 3D rendering TFG while preserving the visual fidelity of protected portrait videos.
Moreover, ErodedFace consistently maintains its defense effectiveness under common transformations and recent purification methods, including JPEG compression~\cite{sandoval_segura2023jpeg}, DiffPure~\cite{nie2022diffpure}, and FreqPure~\cite{pei2025freqpure}.
ErodedFace further achieves a 10$\times$ speedup in defense generation over existing image-based defenses, demonstrating its efficiency and practicality for real-world portrait video protection.

In summary, our main contributions are three-fold:
\begin{itemize}
    \item We reveal the fundamental limitation of directly applying reference-image-conditioned proactive defenses to video-based 3D rendering TFG.
    Existing image-based defenses suffer from a paradigm mismatch, as their adversarial perturbation effects cannot be reliably preserved through personalized 3D representation reconstruction and neural rendering.

    \item We propose \textbf{ErodedFace}, to the best of our knowledge, the first proactive defense framework specifically designed for portrait video protection against unauthorized video-based 3D rendering TFG.
    ErodedFace protects portrait videos by disrupting the reference-video preprocessing phase.

    \item We introduce a curriculum-learning-driven two-stage defense strategy that combines shared-parameter multi-scale disruption with artifact-suppressed robust refinement.
    Extensive experiments demonstrate that ErodedFace achieves effective protection, preserves high visual fidelity, exhibits strong robustness against purification methods, and delivers a 10.0$\times$ speedup over existing image-based defenses.
\end{itemize}

\section{Related Work}
\label{sec:related}

\subsection{Video-based 3D Rendering Talking Face Generation}
Talking Face Generation (TFG) aims to synthesize realistic speaker videos from driving signals such as speech audio.
Early methods mainly relied on GAN-based generation frameworks to animate static reference portraits, but often struggled to preserve identity consistency and natural facial dynamics~\cite{chen2019M-LMD,chen2018_lipGan,prajwal2020_WAV2LIP}.
With the progress of 3D reconstruction and neural rendering, 3D TFG based on portrait videos has become a dominant paradigm for personalized talking face generation.
NeRF-based methods~\cite{guo2021adnerf,li2023ernerf,Athar_2022_RigNeRF,Ye_2024_geneface,Peng_2024_synctalk} learn person-specific radiance fields from reference portrait videos, while 3DGS-based methods~\cite{cho2024gaussiantalker,Li_2024_talkingaussian,Li_2025_instag,peng2025synctalkplusplus} further improve rendering efficiency, deformation modeling, and expression control with explicit Gaussian primitives.
While these advances have enabled increasingly realistic, personalized, and even real-time talking-face video synthesis, they also expose personal portrait videos to severe risks of unauthorized speaker videos forgery and privacy abuse.

\subsection{Proactive Defense against Unauthorized Face Content Generation}
Proactive defense aims to prevent unauthorized face content generation before misuse occurs, typically by injecting visually imperceptible adversarial perturbations into portrait data to degrade downstream forgery results.
Early studies primarily explored proactive protection against GAN-based facial manipulation and DeepFake generation~\cite{huang2021initiative_defense,wanganti_Anti-forgery}.
With the rapid development of diffusion models, subsequent studies have extended proactive defense to diffusion-based facial content generation~\cite{liang2023AdvDM-,salman2023PhotoGuard,xue2023AdvDM+_SDS,jeong2025faceshield}.
Recent efforts have further considered video synthesis scenarios.
I2VGuard~\cite{gui2025i2vguard} protects source images against diffusion-based image-to-video generation by jointly disrupting spatial fidelity, temporal consistency, and denoising behavior.
Silencer~\cite{gan2025silence} protects static portrait images against latent-diffusion-based TFG by nullifying audio control and enhancing anti-purification robustness.
Despite their effectiveness in their respective settings, these defenses remain primarily designed for reference-image-conditioned generation, in which the protected reference image is directly consumed as a conditioning input during forgery.
They do not explicitly against unauthorized video-based 3D rendering TFG, which relies on personalized 3D representation reconstruction and radiance fields rendering rather than direct image-conditioned synthesis.
This paradigm mismatch leaves the privacy risks of unauthorized video-based 3D rendering TFG largely unaddressed.

\section{Method}
\label{sec:method}

\begin{figure}[!t] \centering \includegraphics[width=0.4\textwidth]{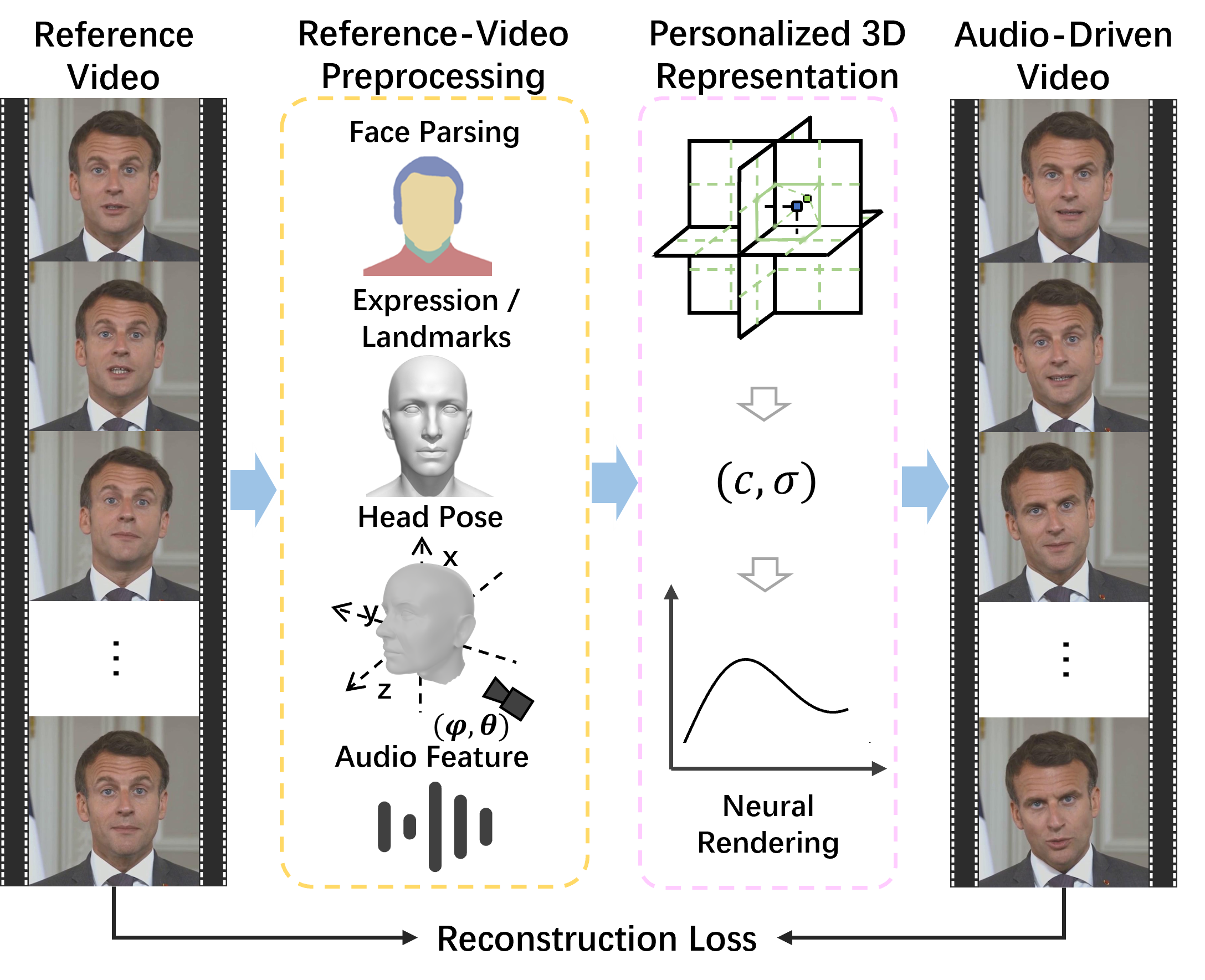} \caption{The basic workflow of the video-based 3D TFG method for acquiring personalized information.} \label{fig:preliminary_pipeline} \end{figure}

\subsection{Paradigm Mismatch and Defense Target}
\label{sec:method_sub_preliminary}

Existing proactive defenses developed for image-based 2D TFG
cannot be directly transferred to video-based 3D rendering TFG because the
protected portrait participates in fundamentally different stages of the two
generation pipelines.
In reference-image-conditioned TFG, the protected portrait remains a direct
visual condition throughout synthesis.
In video-based 3D rendering TFG, however, the reference video is consumed only
during personalization to construct a person-specific 3D representation and
is absent from the subsequent rendering graph.
The two pipelines can be summarized as
\begin{equation}
\label{eq:paradigm_mismatch}
\resizebox{0.88\linewidth}{!}{$\displaystyle
\begin{aligned}
\widehat V_{\mathrm{img}}^{\mathrm{adv}}
&=
G_{\mathrm{img}}
\left(
p+\delta,D_{\mathrm{test}}
\right),
\\
\Phi_{\Delta}^{*}
&=
\mathcal H
\left(
\mathcal P(V+\Delta,A)
\right),
\qquad
\widehat V_{\mathrm{3D}}^{\mathrm{adv}}
=
G_{\mathrm{3D}}
\left(
\Phi_{\Delta}^{*},D_{\mathrm{test}}
\right),
\end{aligned}
$}
\end{equation}
Here, $A$ denotes the audio sequence paired with the reference video during
personalization, whereas $D_{\mathrm{test}}$ denotes the novel driving signals
used during inference.
The image perturbation $\delta$ therefore remains on the deployed synthesis
path, whereas the video perturbation $\Delta$ affects the final output only
indirectly through preprocessing $\mathcal P$ and the optimized representation
$\Phi_{\Delta}^{*}$.
Accordingly, surrogate objectives defined on reference-image encoders, VAEs,
or diffusion denoisers do not directly correspond to the preprocessing and
person-specific optimization modules used by NeRF/3DGS-based TFG.
The relevant defense surface is therefore not the inference-time renderer,
but the upstream interface through which reference-video information is
extracted and accumulated into the personalized representation.

As illustrated in Fig.~\ref{fig:preliminary_pipeline}, face parsing is a
widely adopted and particularly important component of reference-video
preprocessing in video-based 3D rendering TFG~\cite{guo2021adnerf,li2023ernerf,
cho2024gaussiantalker,Li_2024_talkingaussian,Li_2025_instag}.
For each reference frame $I_t$, the face parser $f_{\theta}$ predicts
fine-grained semantic labels, which are merged into three functional regions:
\begin{equation}
\mathbf M_t
=
\mathcal G\!\left(f_{\theta}(I_t)\right)
=
\left\{
M_t^{\mathrm{face}},
M_t^{\mathrm{torso}},
M_t^{\mathrm{bg}}
\right\},
\label{eq:face_parsing_preliminary}
\end{equation}
where $\mathcal G$ denotes the label-grouping operation.
These regions exhibit distinct motion and conditioning characteristics:
the face contains speech-correlated non-rigid dynamics, the torso is dominated
by weaker pose-related motion, and the background is largely static.
Although their concrete implementations differ, existing 3D TFG pipelines use
the resulting masks to route pixels into region-specific representation,
supervision, initialization, or compositing operations.
Persistent parsing errors therefore misroute frame supervision and are
accumulated during multi-frame personalization:
\begin{equation}
\Delta\mathbf M_t
\;\longrightarrow\;
\Delta\mathcal L_{\mathrm{pers}}
\;\longrightarrow\;
\Delta\Phi_{\Delta}^{*}
\;\longrightarrow\;
\Delta\widehat V_{\mathrm{3D}},
\label{eq:parsing_error_propagation}
\end{equation}
where $\mathcal L_{\mathrm{pers}}$ denotes the region-dependent
personalization objective.
Face parsing thus forms a high-leverage upstream bottleneck, and disrupting it can affect different downstream representations without differentiating through their costly person-specific optimization.
Based on this analysis, ErodedFace targets face parsing rather than a
model-specific renderer.
It induces persistent semantic-routing errors in the protected reference
video, causing corrupted supervision to be embedded into the personalized
3D representation before inference begins.
A more detailed formulation of this error propagation is provided in
Appendix~\ref{app:paradigm_mismatch}.

\subsection{ErodedFace}
\label{sec:erodedface}

ErodedFace formulates portrait-video protection as a constrained two-stage
optimization problem.
Directly optimizing parsing disruption, perceptual fidelity, temporal
consistency, and transformation robustness from random initialization
introduces competing objectives.
We therefore decouple defense establishment from fidelity refinement.
Stage I efficiently finds a scale-robust parsing-disruption solution, whereas
Stage II improves its perceptual and temporal quality under explicit
constraints that preserve the established defense.
Conceptually, the two stages solve
\begin{equation}
\label{eq:two_stage_principle}
\resizebox{0.88\linewidth}{!}{$\displaystyle
\begin{aligned}
\text{Stage I:}\quad
&\text{establish }
\min_{s\in\mathcal S} r_{t,s}\geq\tau,
\\
\text{Stage II:}\quad
&\min_{\Theta_t}
\mathcal L_{\mathrm{quality}}(\Theta_t)
\quad
\text{s.t.}\quad
r_{t,s}^{b}\geq\tau,
\ \forall s\in\mathcal S,\ b\in\{0,J\},
\end{aligned}
$}
\end{equation}
where $r_{t,s}^{b}$ is the parsing-erosion rate at scale $s$ under the
raw or JPEG branch.
The following sections instantiate this principle through a shared
frequency-spatial parameterization, an active-scale Stage-I objective,
and a guarded Stage-II refinement objective.

\paragraph{Threat model and overview.}

\paragraph{Threat model.}
Given a reference portrait video
$V=\{I_t\}_{t=1}^{T}$, the defender releases a protected portrait video
$\widetilde V=\{\widetilde I_t\}_{t=1}^{T}$ satisfying
$\|\widetilde I_t-I_t\|_{\infty}\leq\epsilon$.
An adversary may resize, compress, or purify $\widetilde V$ before using it
to personalize a video-based 3D rendering model.
ErodedFace assumes gradient access only to the face-parsing model
$f_{\theta}$ used in reference-video preprocessing.
It neither accesses nor differentiates through the downstream
personalization procedure, learned personalized 3D representation, or
neural renderer.
Our primary evaluation uses the same parser checkpoint for defense
optimization and target preprocessing; thus, ErodedFace is white-box only
with respect to $f_{\theta}$.
At the framework level, any differentiable face-parsing model exposing
compatible facial and background probabilities can instantiate
$f_{\theta}$.

\begin{figure*}[t]
    \centering
    \includegraphics[width=0.85\textwidth]{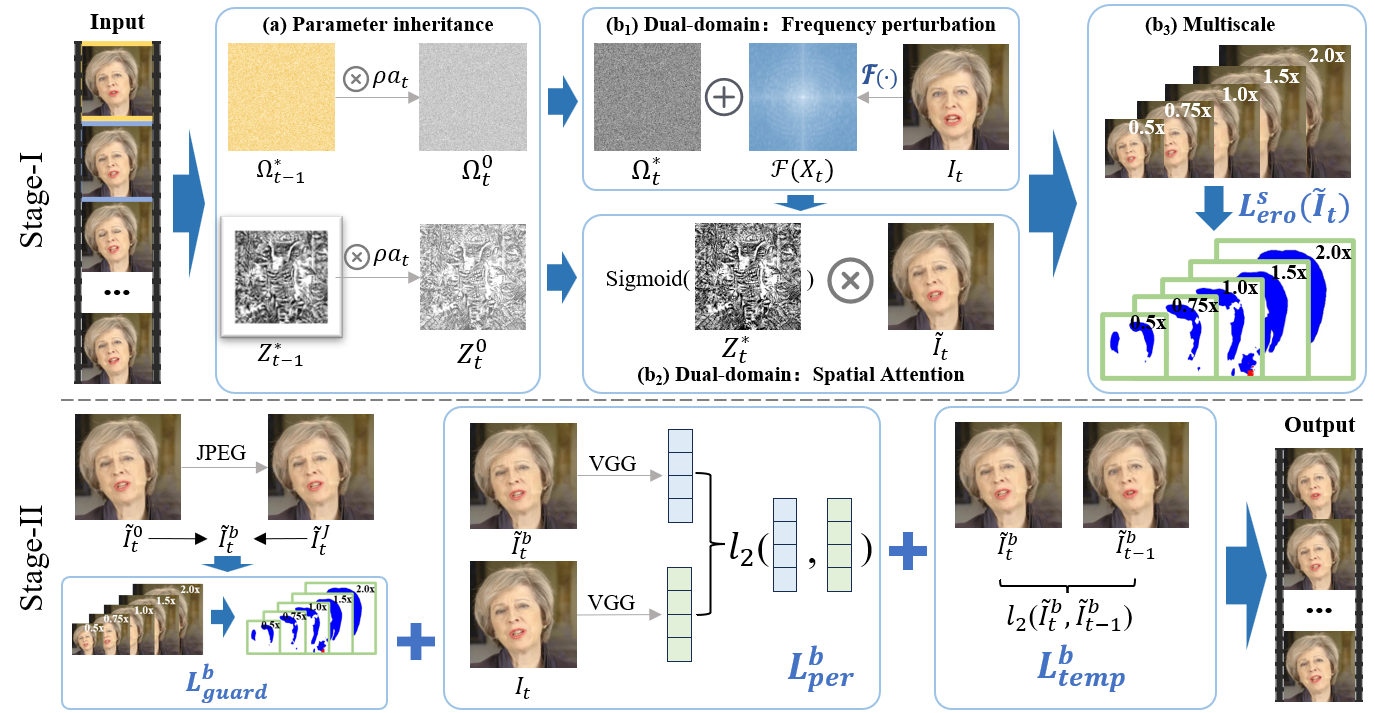}
    \caption{
        \textbf{Overview of ErodedFace.} Stage I initializes the current-frame perturbation parameters through similarity-guided parameter inheritance, applies Fourier amplitude/phase modulation with a learnable spatial gate, and optimizes parsing erosion over the active unresolved-scale set. Stage II refines the protected frame through raw and differentiable-JPEG branches using perceptual and temporal losses, while an active parsing guard preserves the target erosion rate.
    }
    \label{fig:erodedface}
\end{figure*}

As shown in Fig.~\ref{fig:erodedface}, Stage I exploits inter-frame
redundancy through similarity-guided parameter inheritance and establishes
face-to-background erosion over an adaptively updated scale set.
Stage II then refines the Stage-I solution through raw and differentiable-JPEG
branches, using quality objectives to suppress artifacts and active parsing
guards to prevent defense recovery.

\paragraph{Frequency modulation with spatial gating.}
Spectral perturbations typically offer superior robustness against purification and common transformations (such as resizing and compression)~\cite{bala2025dct_shield,jeong2025faceshield}. Therefore, ErodedFace applies perturbations in the frequency domain while eliminating redundant noise in the spatial domain.
To combine cross-scale frequency modulation with spatial selectivity,
for the normalized frame $X_t=\mathcal N(I_t)$, let
$\mathcal F(X_t)=\Gamma_t\exp(\mathrm{i}\Psi_t)$.
The trainable parameters for frame $t$ are
\[
\Theta_t
=
\left(
U_t^{\Gamma},
U_t^{\Psi},
Z_t
\right),
\]
where $U_t^{\Gamma}$ and $U_t^{\Psi}$ modulate Fourier amplitude and phase,
respectively, and $G_t=\sigma(Z_t)$ is a channel-shared spatial gate.
\begin{equation}
\label{eq:protected_frame}
\resizebox{0.88\linewidth}{!}{$\displaystyle
\begin{aligned}
\widehat{\Gamma}_t
&=
\Gamma_t\odot
\left[1+\tanh(U_t^\Gamma)\right],
&
\widehat{\Psi}_t
&=
\Psi_t+g_\Psi(U_t^\Psi),
\\
Y_t
&=
\mathcal F_{\mathbb R}^{-1}
\left(
\widehat{\Gamma}_t e^{\mathrm{i}\widehat{\Psi}_t}
\right),
&
d_t
&=
\epsilon\tanh
\left(
\kappa[\mathcal N^{-1}(Y_t)-I_t]
\right),
\\
\widetilde I_t
&=
Q_{8,\epsilon}
\left(
I_t+\sigma(Z_t)\odot d_t;\,I_t
\right).
\end{aligned}
$}
\end{equation}
Here, $\mathcal F_{\mathbb R}^{-1}$ denotes a real-valued inverse Fourier
transform, and $Q_{8,\epsilon}(\cdot;I_t)$ performs 8-bit quantization while
projecting the result onto the feasible set centered at $I_t$.
The frequency variables determine the adversarial content, whereas the
spatial gate controls where that content is retained.
Exact definitions are provided in Appendix~\ref{app:parameterization}.

\subsubsection{Stage I: Efficient Scale-Robust Parsing Erosion.}
\label{sec:erodedface_stage1}

This stage addresses two video-specific bottlenecks.
First, independently optimizing thousands of highly redundant frames is
computationally wasteful.
Second, a parsing perturbation optimized at one resolution may be weakened
after resizing.
We address these issues through similarity-guided parameter inheritance and
active multi-scale erosion, respectively.

\paragraph{Similarity-guided parameter inheritance.}
Adjacent portrait frames usually exhibit similar identity, facial
structure, appearance, and pose.
Instead of optimizing each frame from a random initialization, ErodedFace
uses the preceding-frame solution as a similarity-dependent warm start:
\begin{equation}
\label{eq:parameter_inheritance}
\resizebox{0.88\linewidth}{!}{$\displaystyle
a_t
=
\operatorname{SSIM}(I_{t-1},I_t),
\
\eta_t
=
\rho a_t,
\
\Theta_t^{(0)}
=
\eta_t\Theta_{t-1}^{*}
+
(1-\eta_t)\Theta_{\mathrm{neu}},
$}
\end{equation}
Here, $\Theta_{\mathrm{neu}}$ denotes the neutral parameter state that
produces no specific perturbation.
Thus, highly similar frames retain most of the preceding solution, whereas
dissimilar frames are initialized closer to the neutral state.
All inherited parameters remain trainable, so inheritance accelerates
optimization without enforcing identical perturbations across frames.
where $*$ denotes optimized preceding-frame parameters and the first frame
is randomly initialized.
The inherited parameters remain fully trainable, enabling adaptation to
the current frame.
Thus, Eq.~\ref{eq:parameter_inheritance} performs parameter inheritance
rather than hard parameter sharing.

\paragraph{Adaptive multi-scale parsing erosion.}
We optimize over
$\mathcal S=\{0.5,0.75,1.0,1.5,2.0\}$.
Let $R_s$ denote differentiable resizing and
$\widehat c_{t,s}(u;J)$ denote the predicted class at location $u$ when
image $J$ is evaluated at scale $s$.
We define the optimization support from the clean frame and keep it fixed
throughout optimization:
\begin{equation}
\label{eq:parsing_erosion_loss}
\resizebox{0.88\linewidth}{!}{$\displaystyle
\begin{aligned}
M_{t,s}
&=
\left\{
u:
\widehat c_{t,s}(u;I_t)
\in\mathcal C_{\mathrm{face}}
\right\},
\\
\mathcal L_{\mathrm{ero}}^{s}(\widetilde I_t)
&=
\frac{1}{|M_{t,s}|}
\sum_{u\in M_{t,s}}
\log
\left[
1-
P_{\theta}
\left(
c_{\mathrm{bg}}
\mid R_s(\widetilde I_t)
\right)_u
+\zeta
\right].
\end{aligned}
$}
\end{equation}
Fixing $M_{t,s}$ prevents the target support from shrinking together with
the corrupted parser prediction, ensuring that optimization continues to
cover the complete original facial region.
This log-complement objective is minimized by increasing the background
probability over the complete clean facial support.
The constant $\zeta$ bounds the objective and prevents numerical instability
as the background probability approaches one.

At iteration $k$, let $r_{t,s}^{(k)}$ denote the proportion of pixels in
$M_{t,s}$ currently classified as background, and define the active
unresolved-scale set
\begin{equation}
\mathcal U_t^{(k)}
=
\left\{
s\in\mathcal S:
r_{t,s}^{(k)}<\tau
\right\}.
\end{equation}
Stage I optimizes
\begin{equation}
\mathcal L_{\mathrm{S1}}^{(k)}
=
\sum_{s\in\mathcal U_t^{(k)}}
w_s
\mathcal L_{\mathrm{ero}}^{s}(\widetilde I_t).
\label{eq:stage1_objective}
\end{equation}
By removing already resolved scales from the active set, Stage I allocates
subsequent updates to the remaining difficult scales.
Optimization stops once all scales satisfy the target erosion rate.
The confirmation anchor and exact stopping criterion are provided in
Appendix~\ref{app:stage1_details}.
Stage I deliberately prioritizes rapid convergence toward the scale-robust defense region without imposing an explicit fidelity objective. As a result, perturbation components inherited from previous frames may persist even when they are no longer necessary for the current frame, leading to artifacts that progressively accumulate and intensify over time. Stage II removes these redundant components while preserving the erosion rates achieved in Stage I as explicit constraints.

\subsubsection{Stage II: Guarded Fidelity and Robustness Refinement}
\label{sec:erodedface_stage2}

Starting from the Stage-I solution, Stage II minimizes perceptual and
temporal artifacts while preserving parsing erosion as an explicit
feasibility constraint.
Unlike re-optimizing protection and fidelity jointly from scratch, this
guarded refinement searches locally around an already effective defense
solution.

We use a raw branch $T_0=\operatorname{Id}$ and a differentiable-JPEG
branch $T_J=\operatorname{DiffJPEG}_q$.
For $b\in\{0,J\}$, we define the branch-matched protected and clean frames as
\begin{equation}
\widetilde I_t^b
=
T_b(\widetilde I_t),
\qquad
I_t^b
=
T_b(I_t).
\label{eq:branch_frames}
\end{equation}
The ideal Stage-II problem is
\begin{equation}
\begin{aligned}
\min_{\Theta_t}\quad
&
\sum_{b\in\{0,J\}}
\lambda_b
\left(
\lambda_{\mathrm{per}}\mathcal L_{\mathrm{per}}^b
+
\lambda_{\mathrm{temp}}\mathcal L_{\mathrm{temp}}^b
\right),
\\
\mathrm{s.t.}\quad
&
r_{t,s}^{b}\geq\tau,
\qquad
\forall b\in\{0,J\},
\ \forall s\in\mathcal S.
\end{aligned}
\label{eq:stage2_constrained}
\end{equation}
We implement these constraints through an active parsing guard that
reactivates the erosion loss only for branches and scales whose erosion rate
falls below $\tau$.
For each branch,
$\mathcal L_{\mathrm{per}}^b$ measures VGG16 feature distance to the clean
frame~\cite{simonyan2015VGG},
$\mathcal L_{\mathrm{temp}}^b$ regularizes inter-frame consistency, and
$\mathcal L_{\mathrm{guard}}^b$ restores parsing erosion at scales that
fall below the target threshold:
\begin{equation}
\label{eq:stage2_objective}
\resizebox{0.88\linewidth}{!}{$\displaystyle
\begin{aligned}
\mathcal L_{\mathrm{guard}}^b
&=
\beta_0
\mathcal L_{\mathrm{ero}}^{1.0}(\widetilde I_t^b)
+
\sum_{s:r_{t,s}^b<\tau}
\mathcal L_{\mathrm{ero}}^{s}(\widetilde I_t^b),
\\
\mathcal L_{\mathrm{S2}}
&=
\sum_{b\in\{0,J\}}
\lambda_b
\left(
\lambda_{\mathrm{per}}
\mathcal L_{\mathrm{per}}^b
+
\lambda_{\mathrm{temp}}
\mathcal L_{\mathrm{temp}}^b
+
\lambda_{\mathrm{guard}}
\mathcal L_{\mathrm{guard}}^b
\right),
\end{aligned}
$}
\end{equation}
where $\lambda_0=1$ and
$\lambda_J=\lambda_{\mathrm{jpeg}}$.
The guard implements an active-set relaxation of the constraints in
Eq.~\ref{eq:stage2_constrained}.
When a branch--scale pair remains feasible, no additional erosion gradient is
introduced; once its erosion rate falls below $\tau$, the corresponding
targeted loss is reactivated.
Stage II can therefore remove redundant perturbation components without
continuously over-optimizing already robust scales.
Stage II terminates when the combined raw- and JPEG-branch perceptual
objective reaches a plateau and all branch-scale erosion constraints remain
satisfied.
Otherwise, optimization continues until the maximum iteration count..
Exact loss definitions, stopping criteria, and hyperparameters are
provided in the Appendix~\ref{app:erodedface_details}.

\section{Experiments}
\label{sec:Exp}

\subsection{Experimental Setup}
\label{sec: Exp setup}
\subsubsection{Implementation Details}
\label{sec:exp_implementation_details}
Following prior 3D talking-face generation methods~\cite{Li_2025_instag,Peng_2024_synctalk}, all videos are resized to 512 \ensuremath{\times} 512 and sampled at 25 FPS, while the audio is resampled to 16 kHz. SyncTalk~\cite{Peng_2024_synctalk} is used as the primary target model, and cross-model evaluations on InsTaG~\cite{Li_2025_instag}, TalkingGaussian~\cite{Li_2024_talkingaussian}, and DEGSTalk~\cite{deng2025degstalk} are reported in Appendix~\ref{sec:transferability}.
The perturbation budget is set to $\epsilon=0.05$ in the normalized $[0,1]$ range.
Optimization terminates when the target-region parsing misclassification rate exceeds 80\% at all five scales.
All experiments are conducted on a single NVIDIA RTX A6000 GPU.

\subsubsection{Baselines and Datasets}
We conduct the main evaluation on 11 portrait videos, including all nine videos from Talk9 and two additional videos adopted from prior work. For each identity, we use a 45-second clip, resulting in 12,375 frames in total. We compare ErodedFace with representative image-level defense methods, including
AdvDM(-)~\cite{liang2023AdvDM-},
PhotoGuard~\cite{salman2023PhotoGuard}, SDS~\cite{xue2023AdvDM+_SDS}, FaceLock~\cite{wang2025facelock}, and Silencer-I~\cite{gan2025silence}, which are applied to the reference videos frame by frame. Further evaluations on HDTF~\cite{zhang2021hdtf}, cross-model transferability to other 3D TFG models, and comparisons with a broader set of image-level defense methods are reported in the Appendix.

\subsubsection{Metrics}
\label{sec:exp_metrics}
We evaluate protected-video fidelity and downstream defense effectiveness using PSNR, SSIM, FID~\cite{heusel2017FID}, Sync-C~\cite{Prajwal_2020_sync_c_d}, and M-LMD~\cite{chen2019M-LMD}. The preferred metric directions for protected and generated videos are marked in the corresponding tables.

\begin{table}[!tbp]
\centering
\resizebox{\linewidth}{!}{
\begin{tabular}{l l| c c c c c c}
\toprule
\multicolumn{2}{c|}{Method} & SSIM$\uparrow$ & PSNR$\uparrow$ & FID$\downarrow$ & Sync-C$\uparrow$ & M-LMD$\downarrow$ & Time(h) \\
\midrule
\multicolumn{2}{l|}{AdvDM(-)\cite{liang2023AdvDM-}} & 0.4793 & 27.48 & 63.40 & 7.8606 & 2.6537 & 22.3 \\
\multicolumn{2}{l|}{PhotoGuard\cite{salman2023PhotoGuard}} & 0.5040 & 26.54 & 107.53 & 7.6222 & 3.1085 & 18.6\\
\multicolumn{2}{l|}{SDS(+)\cite{xue2023AdvDM+_SDS}} & 0.5453 & 27.57 & 127.02 & \textbf{7.9264} & 2.7472 & \underline{12.7}\\
\multicolumn{2}{l|}{SDS(-)\cite{xue2023AdvDM+_SDS}} & 0.5477 & \underline{28.55} & \underline{67.31} & 7.8797 & 2.6503 & 12.8\\
\multicolumn{2}{l|}{SDST(-)\cite{xue2023AdvDM+_SDS}} & \underline{0.5530} & 27.55 & 100.80 & 7.7968 & 2.9659 & 19.5\\
\multicolumn{2}{l|}{Silencer-I\cite{gan2025silence}} & 0.5311 & 28.48 & 78.29 & 7.8652 & \underline{2.6416} & 18.8\\
\midrule
\multicolumn{2}{l|}{ErodedFace (ours)} & \textbf{0.9744} & \textbf{38.69} & \textbf{15.65} & \underline{7.9156} & \textbf{2.3540} & \textbf{1.27}(\textbf{ 10.0$\times$})\\
\midrule
\multicolumn{2}{l|}{Ground Truth} & 1.00 & $\infty$ & 0.00 & 8.0789 & 0.0000 & - \\
\bottomrule
\end{tabular}}
\caption{\textbf{Quantitative Comparisons with representative methods on protected videos.}
Higher SSIM, PSNR, and Sync-C and lower FID and M-LMD indicate better fidelity and audio-visual consistency. Best and second-best results are shown in bold and underlined, respectively. Time is reported for protecting a 45-second video.
}
\label{Protect_Quantitative_Result}
\end{table}

\begin{table}[!b]
\centering
\small
\setlength{\tabcolsep}{4pt}
\resizebox{\linewidth}{!}{
\begin{tabular}{r | c c c c c}
\toprule
\textbf{Method} & \textbf{SSIM$\downarrow$} & \textbf{PSNR$\downarrow$} & \textbf{FID$\uparrow$} & \textbf{Sync-C$\downarrow$} & \textbf{M-LMD$\uparrow$} \\
\midrule
{AdvDM(-)\cite{liang2023AdvDM-}} & 0.8999 & 29.58 & 21.15 & 6.7968 & 9.725\\
{PhotoGuard\cite{salman2023PhotoGuard}} & 0.8890 & 27.58 & 39.67 & \underline{6.0465} & \underline{16.70}\\
{SDS(+)\cite{xue2023AdvDM+_SDS}} & 0.8864 & 29.01 & \underline{61.50} & 6.8855 & 9.032\\
{SDS(-)\cite{xue2023AdvDM+_SDS}} & 0.8973 & 28.49 & 19.57 & 6.7258 & 12.06\\
{SDST(-)\cite{xue2023AdvDM+_SDS}} & \underline{0.8309} & \underline{25.68} & 61.29 & 6.4674 & 13.15\\
{Silencer-I\cite{gan2025silence}} & 0.8983 & 29.59 & 30.36 & 6.7450 & 10.02\\
\midrule
ErodedFace(ours) & \textbf{0.8052} & \textbf{16.27} & \textbf{203.9} & \textbf{0.4276} & \textbf{176.8} \\
\midrule
Ground Truth & 1.00 & $\infty$ & 0.00 & 8.1076 & 0.0000\\
\bottomrule
\end{tabular}}
\caption{\textbf{Quantitative comparison of downstream protection performance on videos generated by SyncTalk from protected reference videos. }
Lower SSIM, PSNR, and Sync-C and higher FID and M-LMD indicate stronger protection.
}
\label{protect_generted_result}
\end{table}

\subsection{Quality of Protected Videos}
\label{sec: Protected Video Metrics}
A practical portrait video defense should preserve the visual fidelity and audio-visual consistency of the released videos. As shown in Table~\ref{Protect_Quantitative_Result}, ErodedFace substantially outperforms all baselines in protected-video fidelity, achieving the best SSIM, PSNR, and FID. It also obtains the lowest M-LMD and the second-best Sync-C, demonstrating strong preservation of lip motion and audio-visual consistency. Moreover, ErodedFace protects a 45-second video in only 1.27 h,
provides a 10.0$\times$ speedup over the fastest baseline in the main comparison. Overall, ErodedFace achieves a compelling balance between fidelity, consistency, and efficiency.

\subsection{Privacy Protection Performance}
\label{sec: Privacy Protection Performance}

We evaluate whether protected reference videos can still support faithful personalized 3D talking-face generation. Specifically, the videos produced by each defense method are used for person-specific SyncTalk training, and the rendered results are compared with those obtained from the corresponding clean references.
As shown in Table~\ref{protect_generted_result}, existing image-level defenses retain relatively high visual similarity and audio-visual synchronization, indicating that their perturbations are largely ineffective at preventing SyncTalk from learning target-specific appearance cues.
In contrast, ErodedFace consistently disrupts rendered appearance, lip motion, and audio-visual synchronization, reducing Sync-C to 0.4276 while increasing FID and M-LMD to 203.9 and 176.8, respectively. These results indicate that ErodedFace prevents SyncTalk from learning a stable and controllable personalized 3D representation.

Cross-model transferability results on InsTaG, TalkingGaussian, and DEGSTalk are reported in Appendix~\ref{sec:transferability}. Additional comparisons with image-level defenses on SyncTalk and evaluations on the HDTF dataset are provided in Appendices~\ref{sec:more defense results} and~\ref{subsec:additional_data}, respectively.

\begin{table}[!htbp]
\centering
\resizebox{0.98\linewidth}{!}{
\begin{tabular}{l l | c c}
\toprule
\multicolumn{2}{c|}{\multirow{2}{*}{Method}}
& JPEG
& Resizing \\
\multicolumn{2}{r|}{}
& PSNR$\downarrow$/FID$\uparrow$/Sync-C$\downarrow$/M-LMD$\uparrow$
& PSNR$\downarrow$/FID$\uparrow$/Sync-C$\downarrow$/M-LMD$\uparrow$ \\
\midrule
\multicolumn{2}{l|}{AdvDM(-)\cite{liang2023AdvDM-}}
& 29.76/17.70/6.9259/9.843
& 30.15/24.16/6.6624/10.26
\\

\multicolumn{2}{l|}{PhotoGuard\cite{salman2023PhotoGuard}}
& \underline{27.14}/71.36/\underline{6.0550}/\underline{24.61}
& 28.23/33.40/\underline{6.1544}/\underline{12.45}
\\

\multicolumn{2}{l|}{SDS(+)\cite{xue2023AdvDM+_SDS}}
& 29.04/55.74/6.5871/9.069
& 29.30/56.96/6.7124/8.859\\

\multicolumn{2}{l|}{SDS(-)\cite{xue2023AdvDM+_SDS}}
& 27.25/62.46/7.1242/22.70
& 29.28/21.69/6.6871/10.18 \\

\multicolumn{2}{l|}{SDST(-)\cite{xue2023AdvDM+_SDS}}
& 27.97/32.16/6.1517/11.88
& \underline{27.95}/31.88/6.3840/12.00 \\

\multicolumn{2}{l|}{Silencer-I\cite{gan2025silence}}
& 30.28/24.36/6.8438/9.016
& 28.66/\underline{60.66}/6.9392/7.121 \\

\midrule
\multicolumn{2}{l|}{ErodedFace(ours)}
&
\textbf{20.79}/\textbf{146.2}/\textbf{2.7437}/\textbf{37.04}
&
\textbf{16.43}/\textbf{187.7}/\textbf{1.2834}/\textbf{94.84}
\\
\bottomrule
\end{tabular}}
\caption{\textbf{Quantitative comparison of downstream protection performance after JPEG compression and resizing. }
The transformed protected videos are used as references for person-specific SyncTalk training. JPEG compression is performed with a quality factor of 75, while resizing downsamples each frame from 512\ensuremath{\times}512 to 400\ensuremath{\times}400 and subsequently restores it to 512\ensuremath{\times}512. The two transformations are evaluated separately.
Lower PSNR and Sync-C and higher FID and M-LMD indicate stronger protection.
}
\label{table3:transformation_analysis}
\end{table}

\subsection{Evaluation under Image Transformations}
\label{sec: Anti-pure exp}

\begin{figure}[!ht]
\centering
\includegraphics[width=0.80\linewidth]{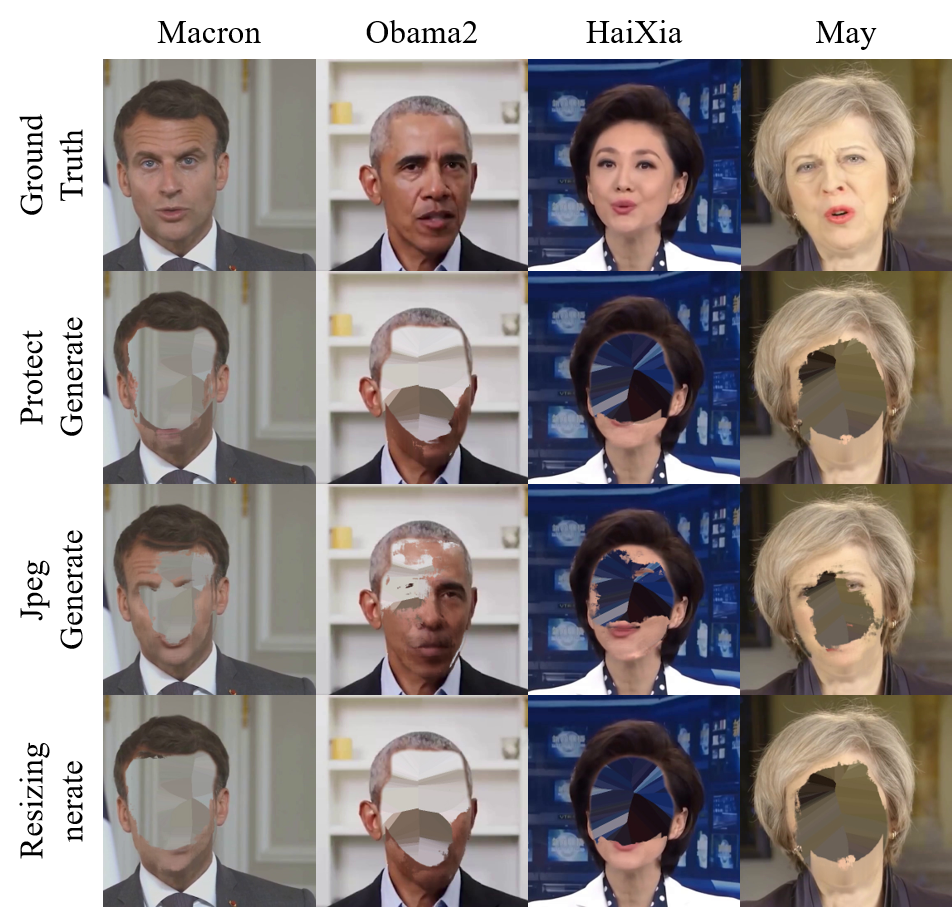}
\caption{\textbf{Qualitative evaluation under JPEG compression and resizing.}
From top to bottom, we show the original reference frame and SyncTalk results obtained using ErodedFace-protected videos without transformation, after JPEG compression, and after resizing, respectively. Facial artifacts and structural distortions persist after both transformations, indicating that the protective effect is retained.}
\label{Qualitative_Comparison_jpeg_resize_gen}
\end{figure}

We evaluate ErodedFace under two types of practical preprocessing operations: common input transformations, including JPEG compression and resizing, and diffusion-based purification methods, including DiffPure~\cite{nie2022diffpure} and FreqPure~\cite{pei2025freqpure}.
\\
\textbf{Robustness to Common Image Transformations.}
As shown in Table~\ref{table3:transformation_analysis}, ErodedFace consistently achieves the strongest protection under both JPEG compression and resizing, outperforming all baselines across the four downstream metrics. In particular, its Sync-C decreases to 2.7437 and 1.2834 under JPEG compression and resizing, respectively, while its FID increases to 146.2 and 187.7.
The qualitative results in Fig.~\ref{Qualitative_Comparison_jpeg_resize_gen} further show that pronounced facial artifacts and structural distortions persist after both transformations.
These results demonstrate that ErodedFace does not rely on a fixed resolution or an uncompressed input representation. Its robustness to resizing is supported by the adaptive multi-scale optimization in Stage I, while the differentiable-JPEG refinement in Stage II improves resistance to compression.

Additional evaluations against DiffPure and FreqPure, neither of which is included in our optimization, are provided in Appendix~\ref{subsec:more anti-purification}.

\subsection{Ablation Study}
\label{sec:ablation}

\begin{table}[!htbp]
\centering
\resizebox{\columnwidth}{!}{
\begin{tabular}{cccc|cccc}
\toprule
\multirow{2}{*}{\textbf{Stage-II}} & \multirow{2}{*}{\textbf{Mask}} & \multirow{2}{*}{\textbf{VGG}} & \multirow{2}{*}{\textbf{Param. Inherit.}} & \multicolumn{4}{c}{\textbf{Protected Image}} \\
\cmidrule(r){5-8} & & & & \textbf{SSIM}$\uparrow$ & \textbf{PSNR}$\uparrow$ & \textbf{FID}$\downarrow$ & \textbf{Time (min)}\\
\midrule
\textcolor{gray}{--} & \textbf{\checkmark} & \textbf{\checkmark} & \textbf{\checkmark} & \underline{0.9613} & \underline{34.97} & 21.08 & 12.91\\
\textbf{\checkmark} & \textcolor{gray}{--} & \textbf{\checkmark} & \textbf{\checkmark} & 0.9549 & 29.10 & \textbf{15.24} & 117.5\\
\textbf{\checkmark} & \textbf{\checkmark} & \textcolor{gray}{--} & \textbf{\checkmark} & 0.9038 & 33.12 & 55.90 & 25.73\\
\textbf{\checkmark} & \textbf{\checkmark} &
\textbf{\checkmark} & \textcolor{gray}{--} & 0.9285 & 34.13 & 24.52 & 297.5\\
\textbf{\checkmark} & \textbf{\checkmark} & \textbf{\checkmark} & \textbf{\checkmark} & \textbf{0.9744} & \textbf{38.69} & \underline{15.65} & 76.18\\
\bottomrule\end{tabular}}
\caption{\textbf{Ablation study of ErodedFace.}
\textcolor{gray}{--} denotes that the corresponding component is removed, while $\checkmark$ denotes that it is enabled. The last row represents the full configuration.}
\label{tab:progressive_ablation}
\end{table}

Table~\ref{tab:progressive_ablation} analyzes the contribution of the main components of ErodedFace. Removing Stage II degrades all fidelity metrics, confirming its role in suppressing residual artifacts. Although removing the spatial mask slightly improves FID, it reduces SSIM and PSNR and increases the runtime, showing the benefit of spatially selective optimization. The VGG loss further improves perceptual quality, while parameter inheritance reduces the runtime from 297.5 to 76.18 min and improves all fidelity metrics. Overall, these components complement each other to achieve a favorable balance between protected-video fidelity and optimization efficiency.

\section{Conclusion}

We identify a mismatch between image-level defenses and video-based 3D TFG: frame-wise perturbations are ineffective because these systems reconstruct a person-specific 3D representation from the reference video rather than directly using protected frames. We therefore propose \textbf{ErodedFace}, a robust portrait-video defense that disrupts reference-video preprocessing and corrupts target-specific geometry and appearance cues. Combining similarity-guided parameter inheritance, frequency modulation, spatial gating, and adaptive multi-scale optimization, ErodedFace provides strong, transformation-robust protection while preserving visual fidelity and improving efficiency. Future work will pursue a unified defense for image- and video-based TFG.

\bibliography{aaai2027}

@inproceedings{prajwal2020_WAV2LIP,
  author    = {Prajwal, K. R. and Mukhopadhyay, Rudrabha and Namboodiri, Vinay P. and Jawahar, C. V.},
  title     = {A Lip Sync Expert Is All You Need for Speech to Lip Generation In the Wild},
  booktitle = {Proceedings of the 28th ACM International Conference on Multimedia (ACM MM)},
  pages     = {484--492},
  year      = {2020},
  location  = {Seattle, WA, USA},
  publisher = {Association for Computing Machinery},
  doi       = {10.1145/3394171.3413532}
}

@inproceedings{chen2018_lipGan,
  author    = {Chen, Lele and Li, Zhiheng and Maddox, Ross K. and Duan, Zhiyao and Xu, Chenliang},
  title     = {Lip Movements Generation at a Glance},
  booktitle = {Proceedings of the European Conference on Computer Vision (ECCV)},
  pages     = {520--535},
  year      = {2018},
  location  = {Munich, Germany},
  publisher = {Springer},
  doi       = {10.1007/978-3-030-01234-2_32}
}

@inproceedings{guo2021adnerf,
  title     = {AD-NeRF: Audio Driven Neural Radiance Fields for Talking Head Synthesis},
  author    = {Guo, Yudong and Chen, Keyu and Liang, Sen and Liu, Yong-Jin and Bao, Hujun and Zhang, Juyong},
  booktitle = {Proceedings of the IEEE/CVF International Conference on Computer Vision (ICCV)},
  year      = {2021},
  pages     = {10104--10113},
  publisher = {IEEE},
  doi       = {10.1109/ICCV48922.2021.00991}
}

@inproceedings{cho2024gaussiantalker,
  author    = {Cho, Kyusun and Lee, Joungbin and Yoon, Heeji and Hong, Yeobin and Ko, Jaehoon and Ahn, Sangjun and Kim, Seungryong},
  title     = {GaussianTalker: Real-Time High-Fidelity Talking Head Synthesis with Audio-Driven 3D Gaussian Splatting},
  booktitle = {Proceedings of the 32nd ACM International Conference on Multimedia (ACM MM)},
  pages     = {10985--10994},
  year      = {2024},
  publisher = {ACM},
  doi       = {10.1145/3664647.3681627}
}

@inproceedings{Peng_2024_synctalk,
  author    = {Peng, Ziqiao and Hu, Wentao and Shi, Yue and Zhu, Xiangyu and Zhang, Xiaomei and Zhao, Hao and He, Jun and Liu, Hongyan and Fan, Zhaoxin},
  title     = {SyncTalk: The Devil is in the Synchronization for Talking Head Synthesis},
  booktitle = {Proceedings of the IEEE/CVF Conference on Computer Vision and Pattern Recognition (CVPR)},
  pages     = {666--676},
  year      = {2024},
  publisher = {IEEE},
  doi       = {10.1109/CVPR52733.2024.00070},
  address   = {Seattle, WA},
}

@inproceedings{Athar_2022_RigNeRF,
  author    = {Athar, ShahRukh and Xu, Zexiang and Sunkavalli, Kalyan and Shechtman, Eli and Shu, Zhixin},
  title     = {RigNeRF: Fully Controllable Neural 3D Portraits},
  booktitle = {Proceedings of the IEEE/CVF Conference on Computer Vision and Pattern Recognition (CVPR)},
  pages     = {20364--20373},
  year      = {2022},
  publisher = {IEEE},
  doi       = {10.1109/CVPR52688.2022.01972},
  address   = {New Orleans, LA},
}

@inproceedings{Gu_2022_styleNerf,
  author    = {Gu, Jiatao and Liu, Lingjie and Wang, Peng and Theobalt, Christian},
  title     = {StyleNeRF: A Style-based 3D-Aware Generator for High-resolution Image Synthesis},
  booktitle = {Proceedings of the International Conference on Learning Representations (ICLR)},
  year      = {2022},
  url       = {https://openreview.net/forum?id=iUuzzTMUw9K},
}

@inproceedings{Prajwal_2020_sync_c_d,
  author    = {Prajwal, K. R. and Mukhopadhyay, R. and Namboodiri, V. P. and Jawahar, C. V.},
  title     = {A Lip Sync Expert Is All You Need for Speech to Lip Generation in the Wild},
  booktitle = {Proceedings of the 28th ACM International Conference on Multimedia (ACM MM)},
  pages     = {484--492},
  year      = {2020},
  publisher = {ACM},
  doi       = {10.1145/3394171.3413532},
  address   = {Seattle, WA},
}

@inproceedings{Ye_2024_geneface,
  author    = {Ye, Zhenhui and Jiang, Ziyue and Ren, Yi and Liu, Jinglin and He, Jinzheng and Zhao, Zhou},
  title     = {GeneFace: Generalized and High-Fidelity Audio-Driven 3D Talking Face Synthesis},
  booktitle = {Proceedings of the Eleventh International Conference on Learning Representations (ICLR)},
  year      = {2024},
  url       = {https://arxiv.org/abs/2301.13430},
}

@inproceedings{Li_2025_instag,
  author    = {Li, Jiahe and Zhang, Jiawei and Bai, Xiao and Zheng, Jin and Zhou, Jun and Gu, Lin},
  title     = {InsTaG: Learning Personalized 3D Talking Head from Few-Second Video},
  booktitle = {Proceedings of the IEEE/CVF Conference on Computer Vision and Pattern Recognition (CVPR)},
  pages     = {10690--10700},
  year      = {2025},
  publisher = {IEEE},
  doi       = {10.1109/CVPR42600.2025.01047},
  address   = {Los Alamitos, CA},
}

@inproceedings{Li_2024_talkingaussian,
  author    = {Li, Jiahe and Zhang, Jiawei and Bai, Xiao and Zheng, Jin and Ning, Xin and Zhou, Jun and Gu, Lin},
  title     = {TalkingGaussian: Structure-Persistent 3D Talking Head Synthesis via Gaussian Splatting},
  booktitle = {European Conference on Computer Vision (ECCV)},
  pages     = {127--145},
  year      = {2024},
  publisher = {Springer Nature Switzerland},
  doi       = {10.1007/978-3-031-72684-2_8}
}

@inproceedings{li2023ernerf,
  author    = {Li, Jiahe and Zhang, Jiawei and Bai, Xiao and Zhou, Jun and Gu, Lin},
  title     = {Efficient Region-Aware Neural Radiance Fields for High-Fidelity Talking Portrait Synthesis},
  booktitle = {Proceedings of the IEEE/CVF International Conference on Computer Vision (ICCV)},
  year      = {2023},
  pages     = {7568--7578},
  publisher = {IEEE},
  doi       = {10.1109/ICCV48922.2023.00743}
}

@inproceedings{liang2023AdvDM-,
  author    = {Liang, Chumeng and Wu, Xiaoyu and Hua, Yang and Zhang, Jiaru and Xue, Yiming and Song, Tao and Xue, Zhengui and Ma, Ruhui and Guan, Haibing},
  title     = {Adversarial Example Does Good: Preventing Painting Imitation from Diffusion Models via Adversarial Examples},
  booktitle = {Proceedings of the 40th International Conference on Machine Learning (ICML)},
  pages     = {20763--20786},
  year      = {2023},
  series    = {Proceedings of Machine Learning Research},
  volume    = {202},
  publisher = {PMLR}
}

@inproceedings{salman2023PhotoGuard,
  title={Raising the cost of malicious AI-powered image editing},
  author={Salman, Hadi and Khaddaj, Alaa and Leclerc, Guillaume and Ilyas, Andrew and M{\k{a}}dry, Aleksander},
  booktitle={Proceedings of the 40th International Conference on Machine Learning},
  pages={29894--29918},
  year={2023}
}

@article{liang2023Mist,
  author    = {Liang, Chumeng and Wu, Xiaoyu},
  title     = {Mist: Towards Improved Adversarial Examples for Diffusion Models},
  journal   = {arXiv preprint arXiv:2305.12683},
  year      = {2023},
}

@inproceedings{gan2025silence,
  author    = {Gan, Yuan and Miao, Jiaxu and Wang, Yunze and Yang, Yi},
  title     = {Silence is Golden: Leveraging Adversarial Examples to Nullify Audio Control in LDM-based Talking-Head Generation},
  booktitle = {Proceedings of the IEEE/CVF Conference on Computer Vision and Pattern Recognition (CVPR)},
  year      = {2025}
}

@article{sandoval_segura2023jpeg,
  author    = {Sandoval-Segura, Pedro and Geiping, Jonas and Goldstein, Tom},
  title     = {JPEG Compressed Images Can Bypass Protections Against AI Editing},
  journal   = {arXiv preprint arXiv:2304.02234},
  year      = {2023},
}

@inproceedings{nie2022diffpure,
  author    = {Nie, Weili and Guo, Brandon and Huang, Yujia and Xiao, Chaowei and Vahdat, Arash and Anandkumar, Animashree},
  title     = {Diffusion Models for Adversarial Purification},
  booktitle = {Proceedings of the 39th International Conference on Machine Learning (ICML)},
  pages     = {16805--16827},
  year      = {2022},
  series    = {Proceedings of Machine Learning Research},
  volume    = {162},
  publisher = {PMLR}
}

@article{pei2025freqpure,
  author    = {Pei, Gaozheng and Ma, Ke and Sun, Yingfei and Xu, Qianqian and Huang, Qingming},
  title     = {Diffusion-based Adversarial Purification from the Perspective of the Frequency Domain},
  journal   = {arXiv preprint arXiv:2505.01267},
  year      = {2025},
}

@article{peng2025synctalkplusplus,
  author    = {Peng, Ziqiao and Hu, Wentao and Ma, Junyuan and Zhu, Xiangyu and Zhang, Xiaomei and Zhao, Hao and Tian, Hui and He, Jun and Liu, Hongyan and Fan, Zhaoxin},
  title     = {SyncTalk++: High-Fidelity and Efficient Synchronized Talking Heads Synthesis Using Gaussian Splatting},
  journal   = {arXiv preprint arXiv:2506.14742},
  year      = {2025},
}

@inproceedings{simonyan2015VGG,
  author    = {Simonyan, Karen and Zisserman, Andrew},
  title     = {Very Deep Convolutional Networks for Large-Scale Image Recognition},
  booktitle = {Proceedings of the International Conference on Learning Representations (ICLR)},
  year      = {2015},
  month     = {May}
}

@inproceedings{heusel2017FID,
  author    = {Heusel, Martin and Ramsauer, Hubert and Unterthiner, Thomas and Nessler, Bernhard and Hochreiter, Sepp},
  title     = {GANs Trained by a Two Time-Scale Update Rule Converge to a Local Nash Equilibrium},
  booktitle = {Advances in Neural Information Processing Systems (NeurIPS)},
  volume    = {30},
  year      = {2017}
}

@inproceedings{chen2019M-LMD,
  author    = {Chen, Lele and Maddox, Ross K. and Duan, Zhiyao and Xu, Chenliang},
  title     = {Hierarchical Cross-Modal Talking Face Generation with Dynamic Pixel-Wise Loss},
  booktitle = {Proceedings of the IEEE/CVF Conference on Computer Vision and Pattern Recognition (CVPR)},
  pages     = {7832--7841},
  year      = {2019}
}

@inproceedings{jeong2025faceshield,
  author    = {Jeong, Jaehwan and Lee, Seunghyun and Kim, Jaeho and Park, Minjung and Choi, Junghyun and Kim, Sungmin},
  title     = {FaceShield: Defending Facial Image Against Deepfake Threats},
  booktitle = {Proceedings of the IEEE/CVF International Conference on Computer Vision (ICCV)},
  year      = {2025}
}

@article{wanganti_Anti-forgery,
   title={Anti-forgery: Towards a stealthy and robust deepfake disruption attack via adversarial perceptual-aware perturbations},
  author={Wang, Run and Huang, Ziheng and Chen, Zhikai and Liu, Li and Chen, Jing and Wang, Lina},
  journal={arXiv preprint arXiv:2206.00477},
  year={2022}
}

@inproceedings{gui2025i2vguard,
  title     = {I2VGuard: Safeguarding Images against Misuse in Diffusion-based Image-to-Video Models},
  author    = {Gui, Dongnan and Guo, Xun and Zhou, Wengang and Lu, Yan},
  booktitle = {Proceedings of the IEEE/CVF Conference on Computer Vision and Pattern Recognition},
  pages     = {12595--12604},
  year      = {2025}
}

@inproceedings{zhang2021hdtf,
  title={Flow-guided one-shot talking face generation with a high-resolution audio-visual dataset},
  author={Zhang, Zhimeng and Li, Lincheng and Ding, Yu and Fan, Changjie},
  booktitle={Proceedings of the IEEE/CVF conference on computer vision and pattern recognition},
  pages={3661--3670},
  year={2021}
}

@inproceedings{huang2021initiative_defense,
  title={Initiative defense against facial manipulation},
  author={Huang, Qidong and Zhang, Jie and Zhou, Wenbo and Zhang, Weiming and Yu, Nenghai},
  booktitle={Proceedings of the AAAI conference on artificial intelligence},
  volume={35},
  number={2},
  pages={1619--1627},
  year={2021}
}

@inproceedings{deng2025degstalk,
  title={Degstalk: decomposed per-embedding gaussian fields for hair-preserving talking face synthesis},
  author={Deng, Kaijun and Zheng, Dezhi and Xie, Jindong and Wang, Jinbao and Xie, Weicheng and Shen, Linlin and Song, Siyang},
  booktitle={ICASSP 2025-2025 IEEE International Conference on Acoustics, Speech and Signal Processing (ICASSP)},
  pages={1--5},
  year={2025},
  organization={IEEE}
}

@inproceedings{wang2025facelock,
  title={Edit away and my face will not stay: Personal biometric defense against malicious generative editing},
  author={Wang, Hanhui and Zhang, Yihua and Bai, Ruizheng and Zhao, Yue and Liu, Sijia and Tu, Zhengzhong},
  booktitle={Proceedings of the Computer Vision and Pattern Recognition Conference},
  pages={23806--23816},
  year={2025}
}

@inproceedings{xue2023AdvDM+_SDS,
  title={Toward effective protection against diffusion-based mimicry through score distillation},
  author={Xue, Haotian and Liang, Chumeng and Wu, Xiaoyu and Chen, Yongxin},
  booktitle={International Conference on Learning Representations},
  volume={2024},
  pages={25734--25758},
  year={2024}
}

@inproceedings{bala2025dct_shield,
  title={Dct-shield: A robust frequency domain defense against malicious image editing},
  author={Bala, Aniruddha and Chowdhury, Rohit and Jaiswal, Rohan and Roheda, Siddharth},
  booktitle={Proceedings of the IEEE/CVF International Conference on Computer Vision},
  pages={18876--18884},
  year={2025}
}
\appendix
\clearpage

\section{Detailed Analysis of the Paradigm Mismatch}
\label{app:paradigm_mismatch}

\subsection{Direct Visual Conditioning in Image-based 2D TFG}
\label{app:image_based_defense}

Let $p$ denote a reference portrait and $D_{\mathrm{test}}$ denote the
driving signals used for synthesis, such as speech audio.
An image-based 2D TFG model directly consumes the reference
portrait during inference:
\begin{equation}
\widehat{V}_{\mathrm{img}}
=
G_{\mathrm{img}}
\left(
p,D_{\mathrm{test}}
\right).
\label{eq:app_image_generation}
\end{equation}
A proactive defense releases a protected portrait
$p^{\mathrm{adv}}=p+\delta$ under an imperceptibility constraint:
\begin{equation}
\widehat{V}_{\mathrm{img}}^{\mathrm{adv}}
=
G_{\mathrm{img}}
\left(
p+\delta,D_{\mathrm{test}}
\right),
\qquad
\|\delta\|_{\infty}\leq\epsilon.
\label{eq:app_image_adv_generation}
\end{equation}
A representative output-level protection objective can be expressed as
\begin{equation}
\label{eq:app_image_defense}
\resizebox{0.88\linewidth}{!}{$\displaystyle
\delta^{*}
=
\arg\max_{\|\delta\|_{\infty}\leq\epsilon}
\mathcal{L}_{\mathrm{img}}
\left(
G_{\mathrm{img}}(p+\delta,D_{\mathrm{test}}),
G_{\mathrm{img}}(p,D_{\mathrm{test}})
\right),
$}
\end{equation}
where $\mathcal{L}_{\mathrm{img}}$ measures the discrepancy between
generations obtained from protected and clean portraits.
Depending on the target pipeline, this discrepancy may characterize
identity distortion, generation degradation, temporal inconsistency, or
audio-motion mismatch.

Other image-level defenses instantiate the same principle through
surrogate objectives defined on a reference-image encoder, VAE, or
diffusion denoising module.
Despite their different optimization targets, these methods share the
following direct dependency:
\begin{equation}
p^{\mathrm{adv}}
\longrightarrow
G_{\mathrm{img}}
\longrightarrow
\widehat{V}_{\mathrm{img}}^{\mathrm{adv}}.
\label{eq:app_direct_path}
\end{equation}
The protected portrait therefore remains on the deployed synthesis path,
allowing its perturbation to directly affect inference without
differentiating through a person-specific optimization procedure.

\subsection{Personalization in Video-based 3D TFG}
\label{app:video_based_personalization}

Given a reference portrait video
$V=\{I_t\}_{t=1}^{T}$ and its temporally aligned audio sequence
$A=\{a_t\}_{t=1}^{T}$, video-based 3D TFG first performs
person-specific optimization to construct a personalized 3D
representation.
For each reference frame, the preprocessing pipeline extracts audio,
geometric, and semantic cues:
\begin{equation}
\mathbf{c}_t
=
\left[
\mathcal{E}_{a}(a_t),
\mathcal{E}_{g}(I_t),
f_{\theta}(I_t)
\right],
\label{eq:app_conditioning_cues}
\end{equation}
where $\mathcal{E}_{a}$ denotes audio feature extraction,
$\mathcal{E}_{g}$ denotes geometric cue extraction, such as camera pose,
facial expression, or facial landmarks, and $f_{\theta}$ denotes the face
parser used during reference-video preprocessing.

A generic person-specific representation learning objective can be
written as
\begin{equation}
\Phi^{*}
=
\arg\min_{\Phi}
\sum_{t=1}^{T}
\mathcal{L}_{\mathrm{rec}}
\left(
\mathcal{R}_{\Phi}(\mathbf{c}_t),
I_t
\right)
+
\lambda_{\mathrm{reg}}
\mathcal{L}_{\mathrm{reg}}(\Phi),
\label{eq:app_3d_representation}
\end{equation}
where $\mathcal{R}_{\Phi}$ denotes NeRF-based volume rendering or
3DGS-based Gaussian rasterization.
The parameters $\Phi$ collectively describe the person-specific
appearance, geometry, deformation, and rendering representation.

Once $\Phi^{*}$ has been learned, synthesis is performed as
\begin{equation}
\widehat{V}_{\mathrm{3D}}
=
G_{\mathrm{3D}}
\left(
\Phi^{*},D_{\mathrm{test}}
\right).
\label{eq:app_3d_inference}
\end{equation}
The original reference video is absent from this deployed synthesis
graph.
With $\Phi^{*}$ fixed, gradients from the generated video do not directly
propagate to the reference frames.
Accounting for the complete dependency on $V$ would instead require
differentiating through the person-specific optimization that produces
$\Phi^{*}$, resulting in a costly bilevel problem that is fundamentally
different from attacking a direct inference-time visual condition.

\subsection{Why Frame-wise Image Defenses Do Not Directly Transfer}
\label{app:framewise_mismatch}

A straightforward extension of an image-level defense independently
protects every reference frame:
\begin{equation}
V^{\mathrm{adv}}
=
V+\Delta,
\qquad
\Delta=\{\delta_t\}_{t=1}^{T},
\qquad
\|\Delta\|_{\infty}\leq\epsilon.
\label{eq:app_protected_video}
\end{equation}
The unauthorized model then learns a personalized representation from the
protected video:
\begin{equation}
\label{eq:app_adv_personalization}
\resizebox{0.88\linewidth}{!}{$\displaystyle
\Phi_{\Delta}^{*}
=
\arg\min_{\Phi}
\sum_{t=1}^{T}
\mathcal{L}_{\mathrm{rec}}
\left(
\mathcal{R}_{\Phi}
\left(
\mathbf{c}_{t}^{\mathrm{adv}}
\right),
I_t+\delta_t
\right)
+
\lambda_{\mathrm{reg}}
\mathcal{L}_{\mathrm{reg}}(\Phi),
$}
\end{equation}
where
\begin{equation}
\mathbf{c}_{t}^{\mathrm{adv}}
=
\left[
\mathcal{E}_{a}(a_t),
\mathcal{E}_{g}(I_t+\delta_t),
f_{\theta}(I_t+\delta_t)
\right].
\label{eq:app_adv_cues}
\end{equation}
The final synthesized video is
\begin{equation}
\widehat{V}_{\mathrm{3D}}^{\mathrm{adv}}
=
G_{\mathrm{3D}}
\left(
\Phi_{\Delta}^{*},D_{\mathrm{test}}
\right).
\label{eq:app_adv_rendering}
\end{equation}

Unlike the perturbation in image-based 2D TFG, $\Delta$ is not directly
provided to $G_{\mathrm{3D}}$.
Its effect is mediated by reference-video preprocessing, multi-frame
aggregation, and the optimized representation $\Phi_{\Delta}^{*}$.
Consequently, perturbations designed for direct image-conditioned
generation can be attenuated or absorbed during person-specific
reconstruction and do not necessarily remain effective after
personalization.

The corresponding surrogate objectives are also misaligned.
Losses designed to disrupt reference-image encoders, VAEs, or diffusion
denoisers cannot be directly instantiated when the target pipeline instead
relies on semantic preprocessing, personalized NeRF or 3DGS optimization,
and neural rendering.
We therefore direct the protection objective toward an upstream component whose errors are repeatedly propagated throughout person-specific representation learning.

\subsection{Face Parsing as an Upstream Modeling Bottleneck}
\label{app:parsing_bottleneck}

For each reference frame, the face parser predicts fine-grained semantic
labels that are grouped into functional regions:
\begin{equation}
\label{eq:app_parsing_masks}
\resizebox{0.88\linewidth}{!}{$\displaystyle
\mathbf{M}_t
=
\mathcal{G}
\left(
f_{\theta}(I_t)
\right)
=
\left\{
M_t^{\mathrm{face}},
M_t^{\mathrm{torso}},
M_t^{\mathrm{bg}}
\right\},
\qquad
\sum_{r\in\mathcal{R}}M_t^{r}
=
\mathbf{1},
$}
\end{equation}
where
$\mathcal{R}=\{\mathrm{face},\mathrm{torso},\mathrm{bg}\}$ and
$\mathcal{G}$ denotes semantic label grouping.

These regions exhibit different motion and conditioning characteristics.
The face contains the primary speech-correlated non-rigid dynamics.
The torso is dominated by weaker pose-related motion, whereas the
background is generally static.
Separating these regions prevents a single representation from
simultaneously explaining incompatible motion patterns.

Although concrete implementations differ, the role of parsing masks in
person-specific optimization can be abstracted as
\begin{equation}
\mathcal{L}_{\mathrm{pers}}
=
\sum_{t=1}^{T}
\sum_{r\in\mathcal{R}}
\lambda_r
\,
\ell
\left(
M_t^{r}\odot
\mathcal{R}_{\Phi_r}(\mathbf{c}_t),
M_t^{r}\odot I_t
\right),
\label{eq:app_region_reconstruction}
\end{equation}
where $\Phi_r$ denotes the representation or rendering branch associated
with region $r$.
This abstraction covers region-specific representation learning,
foreground-background separation, head-torso decomposition, semantic
supervision, Gaussian initialization, and output compositing.

For a protected frame $I_t^{\mathrm{adv}}=I_t+\delta_t$, the semantic
regions become
\begin{equation}
\mathbf{M}_t^{\mathrm{adv}}
=
\mathcal{G}
\left(
f_{\theta}(I_t+\delta_t)
\right).
\label{eq:app_adv_parsing}
\end{equation}
When pixels are routed to incorrect semantic regions, the resulting masks
provide inaccurate supervision for person-specific representation
learning.
The resulting error propagation can be summarized as
\begin{equation}
\Delta\mathbf{M}_t
\longrightarrow
\Delta\mathcal{L}_{\mathrm{pers}}
\longrightarrow
\Delta\Phi_{\Delta}^{*}
\longrightarrow
\Delta\widehat{V}_{\mathrm{3D}}^{\mathrm{adv}}.
\label{eq:app_error_propagation}
\end{equation}
Because the personalized representation aggregates supervision from many
reference frames, persistent semantic-routing errors can be accumulated
throughout person-specific optimization and affect the learned appearance,
geometry, and deformation representation.

A generic parsing-oriented protection objective can therefore be written
as
\begin{equation}
\label{eq:app_parsing_defense}
\resizebox{0.88\linewidth}{!}{$\displaystyle
\Delta^{*}
=
\arg\min_{\|\Delta\|_{\infty}\leq\epsilon}
\sum_{t=1}^{T}
\mathcal{L}_{\mathrm{ero}}
\left(
\mathcal{G}
\left(
f_{\theta}(I_t+\delta_t)
\right)
\right)
+
\lambda_{\mathrm{per}}
\mathcal{L}_{\mathrm{per}}
\left(
V+\Delta,V
\right),
$}
\end{equation}
where $\mathcal{L}_{\mathrm{ero}}$ disrupts semantic routing and
$\mathcal{L}_{\mathrm{per}}$ preserves the perceptual fidelity of the
released video.
ErodedFace instantiates this principle through scale robust parsing
erosion, frequency modulation with spatial gating, similarity guided
parameter inheritance, and guarded fidelity refinement.

\section{Additional Details of ErodedFace}
\label{app:erodedface_details}

\subsection{Exact Frequency-Spatial Parameterization}
\label{app:parameterization}

For each clean frame $I_t\in[0,1]^{3\times H\times W}$, we apply
ImageNet normalization $X_t=\mathcal{N}(I_t)$ and decompose
\begin{equation}
\mathcal{F}(X_t)
=
\Gamma_t\exp(\mathrm{i}\Psi_t),
\label{eq:app_frequency_decomposition}
\end{equation}
where $\Gamma_t$ and $\Psi_t$ denote the Fourier amplitude and phase.

The channel-shared trainable parameters
$U_t^{\Gamma},U_t^{\Psi}\in\mathbb{R}^{H\times W}$ are broadcast over
the three color channels:
\begin{equation}
\widehat{\Gamma}_t
=
\Gamma_t\odot
\left(
1+\tanh(U_t^{\Gamma})
\right),
\qquad
\widehat{\Psi}_t
=
\Psi_t+g_{\Psi}(U_t^{\Psi}),
\label{eq:app_frequency_modulation}
\end{equation}
where the bounded phase modulation is implemented as
\begin{equation}
g_{\Psi}(u)=\tanh(u).
\label{eq:app_phase_mapping}
\end{equation}
The modulated signal is reconstructed as
\begin{equation}
Y_t
=
\operatorname{Re}
\left\{
\mathcal{F}^{-1}
\left(
\widehat{\Gamma}_t
\exp(\mathrm{i}\widehat{\Psi}_t)
\right)
\right\}.
\label{eq:app_frequency_reconstruction}
\end{equation}

The reconstructed signal is converted into an
$\ell_{\infty}$-bounded residual:
\begin{equation}
d_t
=
\epsilon
\tanh
\left(
\kappa
\left[
\mathcal{N}^{-1}(Y_t)-I_t
\right]
\right).
\label{eq:app_bounded_residual}
\end{equation}
A channel-shared learnable spatial gate
$G_t=\operatorname{sigmoid}(Z_t)$ controls where the residual is retained:
\begin{equation}
\widetilde{I}_t
=
\mathcal{Q}_{8,\epsilon}
\left(
I_t+G_t\odot d_t;
I_t
\right).
\label{eq:app_protected_frame}
\end{equation}
Here, $\mathcal{Q}_{8,\epsilon}(\cdot;I_t)$ performs 8-bit quantization
and projects the result onto the feasible set centered at $I_t$.
A straight-through estimator is used for quantization during
back-propagation.

The trainable variables for frame $t$ are
\begin{equation}
\Theta_t
=
\left(
U_t^{\Gamma},
U_t^{\Psi},
Z_t
\right).
\label{eq:app_trainable_parameters}
\end{equation}
Setting $U_t^{\Gamma}=U_t^{\Psi}=0$ leaves the Fourier representation
unchanged and therefore defines the neutral frequency state.
Both optimization stages update $\Theta_t$ using Adam.

\subsection{Detailed Stage-I Optimization}
\label{app:stage1_details}

For the first frame, the perturbation parameters are randomly initialized.
For each subsequent frame, ErodedFace uses the optimized preceding frame
parameters as a similarity dependent initialization:
\begin{equation}
a_t
=
\operatorname{SSIM}(I_{t-1},I_t),
\qquad
\eta_t
=
\rho a_t,
\label{eq:app_similarity_weight}
\end{equation}
\begin{equation}
\Theta_t^{(0)}
=
\eta_t\Theta_{t-1}^{*}
+
(1-\eta_t)\Theta_{\mathrm{neu}},
\label{eq:app_parameter_inheritance}
\end{equation}
where $\Theta_{\mathrm{neu}}$ denotes the neutral parameter state.
All inherited parameters remain trainable.
Equation~\ref{eq:app_parameter_inheritance} therefore provides a warm
start rather than enforcing identical perturbations across adjacent
frames.

For scale $s$, let
\begin{equation}
\label{eq:app_scaled_prediction}
\resizebox{0.88\linewidth}{!}{$\displaystyle
p_{t,s}^{c}(u;J)
=
P_{\theta}
\left(
c\mid\mathcal{R}_{s}(J)
\right)_{u},
\qquad
\widehat{c}_{t,s}(u;J)
=
\arg\max_{c}p_{t,s}^{c}(u;J),
$}
\end{equation}
where $\mathcal{R}_{s}$ denotes differentiable resizing.
We use the scale set
\begin{equation}
\mathcal{S}
=
\{0.5,0.75,1.0,1.5,2.0\}.
\label{eq:app_scale_set}
\end{equation}

For the 19-class CelebAMask-HQ label space used by the BiSeNet parser,
the facial class set is
$\mathcal{C}_{\mathrm{face}}=\{1,\ldots,13\}$ and the background class is
$c_{\mathrm{bg}}$.
At each scale, the optimization support is determined from the clean
frame:
\begin{equation}
\mathcal{M}_{t,s}
=
\left\{
u:
\widehat{c}_{t,s}(u;I_t)
\in
\mathcal{C}_{\mathrm{face}}
\right\}.
\label{eq:app_clean_support}
\end{equation}
This support remains fixed throughout optimization, preventing the target
region from shrinking together with the corrupted parser prediction.

The scale-specific parsing erosion loss is
\begin{equation}
\mathcal{L}_{\mathrm{ero}}^{s}
\left(
\widetilde{I}_t
\right)
=
\frac{1}{|\mathcal{M}_{t,s}|}
\sum_{u\in\mathcal{M}_{t,s}}
\log
\left[
1-
p_{t,s}^{c_{\mathrm{bg}}}
\left(
u;\widetilde{I}_t
\right)
+
\zeta
\right].
\label{eq:app_erosion_loss}
\end{equation}
Minimizing this objective increases the background probability over the
complete facial support of the clean frame.
The constant $\zeta$ prevents numerical instability as the background
probability approaches one.

The face-to-background erosion rate is
\begin{equation}
r_{t,s}
=
\frac{1}{|\mathcal{M}_{t,s}|}
\sum_{u\in\mathcal{M}_{t,s}}
\mathbb{I}
\left[
\widehat{c}_{t,s}
\left(
u;\widetilde{I}_t
\right)
=
c_{\mathrm{bg}}
\right].
\label{eq:app_erosion_rate}
\end{equation}
At iteration $k$, the unresolved-scale set is
\begin{equation}
\mathcal{U}_{t}^{(k)}
=
\left\{
s\in\mathcal{S}:
r_{t,s}^{(k)}<\tau
\right\}.
\label{eq:app_unresolved_scales}
\end{equation}
The Stage-I objective is
\begin{equation}
\mathcal{L}_{\mathrm{S1}}^{(k)}
=
\sum_{s\in\mathcal{U}_{t}^{(k)}}
w_s
\mathcal{L}_{\mathrm{ero}}^{s}
\left(
\widetilde{I}_t
\right)
+
\beta_1
\mathbb{I}
\left[
\mathcal{U}_{t}^{(k)}=\varnothing
\right]
\mathcal{L}_{\mathrm{ero}}^{1.0}
\left(
\widetilde{I}_t
\right).
\label{eq:app_stage1_objective}
\end{equation}
The first term focuses optimization on scales that have not reached the
target erosion rate.
Once all scales are resolved, the small native-scale anchor retains a
nonzero gradient during confirmation.
The active set is recomputed after every iteration, and Stage I terminates
after all scales satisfy $r_{t,s}\geq\tau$ for three consecutive
evaluations.

\subsection{Selection of the Target Erosion Rate}
\label{app:erosion_rate_selection}

We examine
$\tau\in\{0.2,0.4,0.6,0.8\}$ on the complete main evaluation set,
containing 11 portrait videos of 45 seconds each.
When aggregated over all identities, the difference in total optimization
time between the fastest and slowest settings is less than one hour.
This corresponds to an average difference of less than
$5.45$ minutes per identity.

The limited runtime variation is consistent with an offsetting effect
introduced by similarity-guided parameter inheritance.
A lower target erosion rate allows an isolated frame to satisfy the
Stage-I stopping criterion earlier.
However, the resulting parameters provide a weaker initialization for the
following frame and can require additional adaptation when facial
appearance or pose changes.
The single-frame reduction in optimization cost is therefore partly
offset by less effective inter-frame inheritance, resulting in similar
video-level runtimes across the tested settings.

Because the overall runtime is not sensitive to $\tau$ within this range,
we use $\tau=0.8$ to enforce a more complete face-to-background erosion
before the parameters are inherited by the next frame and refined in
Stage II.

\subsection{Detailed Stage-II Optimization}
\label{app:stage2_details}

Stage II uses a raw branch and a differentiable-JPEG branch:
\begin{equation}
\mathcal{T}_{0}
=
\operatorname{Id},
\qquad
\mathcal{T}_{J}
=
\operatorname{DiffJPEG}_{q}.
\label{eq:app_stage2_branches}
\end{equation}
For branch $b\in\{0,J\}$, the protected and clean inputs are transformed
consistently:
\begin{equation}
\widetilde{I}_{t}^{b}
=
\mathcal{T}_{b}
\left(
\widetilde{I}_{t}
\right),
\qquad
I_t^{b}
=
\mathcal{T}_{b}
\left(
I_t
\right).
\label{eq:app_branch_matched_frames}
\end{equation}
Their normalized forms are
\begin{equation}
\overline{\widetilde{I}}_{t}^{b}
=
\mathcal{N}
\left(
\widetilde{I}_{t}^{b}
\right),
\qquad
\overline{I}_{t}^{b}
=
\mathcal{N}
\left(
I_t^{b}
\right).
\label{eq:app_branch_normalization}
\end{equation}

The branch-specific perceptual loss is
\begin{equation}
\mathcal{L}_{\mathrm{per}}^{b}
=
\sum_{\ell\in\mathcal{K}}
\frac{\gamma_{\ell}}{N_{\ell}}
\left\|
\phi_{\ell}
\left(
\overline{\widetilde{I}}_{t}^{b}
\right)
-
\phi_{\ell}
\left(
\overline{I}_{t}^{b}
\right)
\right\|_{2}^{2},
\label{eq:app_perceptual_loss}
\end{equation}
where $\phi_{\ell}$ denotes the activation map of the $\ell$-th selected
layer of a frozen VGG16 network and $N_{\ell}$ denotes its number of
elements.

The temporal consistency loss is
\begin{equation}
\mathcal{L}_{\mathrm{temp}}^{b}
=
\frac{1}{N}
\left\|
\overline{\widetilde{I}}_{t}^{b}
-
\operatorname{sg}
\left(
\overline{\widetilde{I}}_{t-1}^{b}
\right)
\right\|_{2}^{2},
\label{eq:app_temporal_loss}
\end{equation}
where $\operatorname{sg}$ stops gradients through the preceding protected
frame.
We set $\mathcal{L}_{\mathrm{temp}}^{b}=0$ for the first frame.

Let $r_{t,s}^{b}$ denote the erosion rate after applying branch $b$, and
define
\begin{equation}
\mathcal{U}_{t}^{b}
=
\left\{
s\in\mathcal{S}:
r_{t,s}^{b}<\tau
\right\}.
\label{eq:app_stage2_unresolved_scales}
\end{equation}
The branch-specific parsing guard is
\begin{equation}
\mathcal{L}_{\mathrm{guard}}^{b}
=
\beta_0
\mathcal{L}_{\mathrm{ero}}^{1.0}
\left(
\widetilde{I}_{t}^{b}
\right)
+
\sum_{s\in\mathcal{U}_{t}^{b}}
\mathcal{L}_{\mathrm{ero}}^{s}
\left(
\widetilde{I}_{t}^{b}
\right).
\label{eq:app_parsing_guard}
\end{equation}
The native-scale anchor prevents complete removal of the established
protection, whereas the active terms restore only branch-scale pairs that
fall below the target erosion rate.

The branch-specific refinement objective is
\begin{equation}
\mathcal{L}_{\mathrm{ref}}^{b}
=
\lambda_{\mathrm{per}}
\mathcal{L}_{\mathrm{per}}^{b}
+
\lambda_{\mathrm{temp}}
\mathcal{L}_{\mathrm{temp}}^{b}
+
\lambda_{\mathrm{guard}}
\mathcal{L}_{\mathrm{guard}}^{b}.
\label{eq:app_branch_refinement}
\end{equation}
The complete Stage-II objective is
\begin{equation}
\mathcal{L}_{\mathrm{S2}}
=
\mathcal{L}_{\mathrm{ref}}^{0}
+
\lambda_{\mathrm{jpeg}}
\mathcal{L}_{\mathrm{ref}}^{J}.
\label{eq:app_stage2_objective}
\end{equation}

Stage II terminates when the combined raw-branch and JPEG-branch
perceptual objective reaches a plateau and all branch-scale erosion
constraints remain satisfied.
The plateau is detected using an eight-iteration normalized slope window
with a tolerance of $0.008$.
Otherwise, optimization continues until the maximum iteration count.

\subsection{Implementation Details}
\label{app:implementation_details}

We set
\begin{equation}
\label{eq:app_basic_hyperparameters}
\resizebox{0.88\linewidth}{!}{$\displaystyle
\epsilon=0.05,
\qquad
\kappa=10,
\qquad
\rho=0.8,
\qquad
\tau=0.8,
\qquad
\zeta=10^{-4}.
$}
\end{equation}
The scale weights are
\begin{equation}
\label{eq:app_scale_weights}
\resizebox{0.88\linewidth}{!}{$\displaystyle
w_{1.0}=1.0,
\qquad
w_{1.5}=0.8,
\qquad
w_s=0.6
\quad
\text{for }
s\in\{0.5,0.75,2.0\}.
$}
\end{equation}

Let
$\Omega_t=(U_t^{\Gamma},U_t^{\Psi})$ denote the frequency parameters.
Stage I uses learning rates
$(5\times10^{-2},10^{-2})$ for $(\Omega_t,Z_t)$.
Stage II uses a learning rate of $10^{-2}$ for both parameter groups.
Each stage performs at most 120 iterations per frame.

The loss weights are
\begin{equation}
\left(
\lambda_{\mathrm{per}},
\lambda_{\mathrm{guard}},
\lambda_{\mathrm{temp}},
\lambda_{\mathrm{jpeg}}
\right)
=
(2,0.3,1,1),
\label{eq:app_loss_weights}
\end{equation}
with
\begin{equation}
\beta_0=\beta_1=0.01,
\qquad
q=75.
\label{eq:app_anchor_weights}
\end{equation}
The selected VGG16 layers are
$\mathcal{K}=\{3,8,15,22\}$ with corresponding weights
$(1.0,0.75,0.5,0.25)$.
Parsing success and unresolved scales are evaluated after every
optimization iteration.
\section{Cross Model Transferability Across Video-based 3D TFG methods}
\label{sec:transferability}

\begin{table}[!htbp]
\centering
\small
\setlength{\tabcolsep}{6pt}
\resizebox{\linewidth}{!}{
\begin{tabular}{l|ccccc}
\toprule
\textbf{Method} & \textbf{SSIM$\downarrow$} & \textbf{PSNR$\downarrow$} & \textbf{FID$\uparrow$} & \textbf{Sync-C$\downarrow$} & \textbf{M-LMD$\uparrow$} \\
\midrule
DEGSTalk~\cite{deng2025degstalk} & 0.6984 & 18.96    & 46.62 & 1.0610 &  36.84\\
TalkingGaussian~\cite{Li_2024_talkingaussian} & 0.6859 & 18.62  & 139.2 & 1.0995 &  47.95\\
InsTaG~\cite{Li_2025_instag}    & 0.7913 & 17.11  & 159.4 & 1.5457 & 61.21 \\
\midrule
Ground Truth      & 1.0000   & $\infty$ & 0.000 & 8.0789 & 0.000  \\
\bottomrule
\end{tabular}}
\caption{\textbf{Cross-model transferability of ErodedFace across video-based 3D TFG methods.}
The same ErodedFace protected reference videos are used to personalize each target model without re-optimization or model-specific adaptation, and all models are rendered using the same driving audio. Metrics are evaluated on the resulting generated videos. Lower SSIM, PSNR, and Sync-C and higher FID and M-LMD indicate stronger protection.}
\label{tab:additional_method_performance}
\end{table}

To determine whether ErodedFace depends on a particular personalized
renderer, we generate each protected reference video once using the same
face parser adopted in the primary SyncTalk evaluation.
The protected videos are then used without modification to personalize
InsTaG, TalkingGaussian, and DEGSTalk.
No target specific protection optimization or renderer specific
adaptation is performed, and all methods are evaluated using the same
driving audio.

\begin{figure}[!t]
\centering
\includegraphics[width=\linewidth]{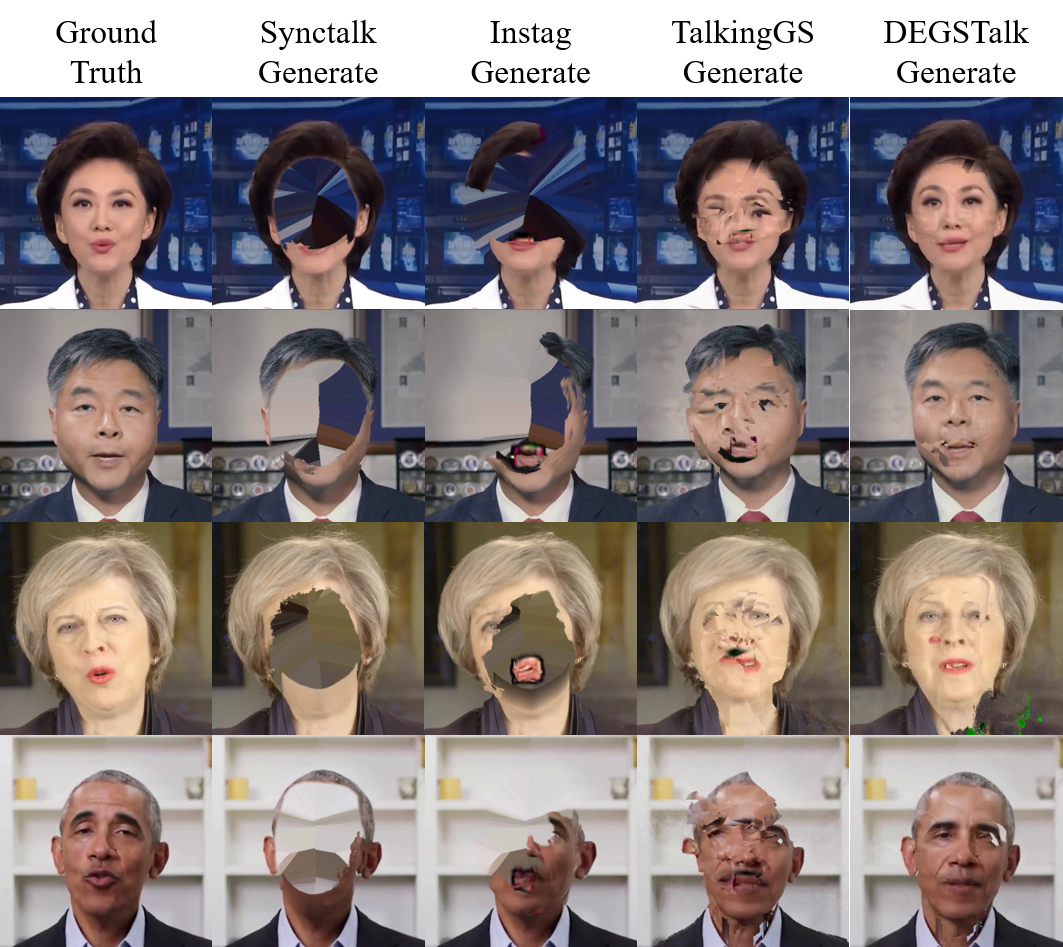}
\caption{\textbf{Qualitative cross model transferability of ErodedFace.}
Generated frames from SyncTalk, InsTaG, TalkingGaussian, and DEGSTalk exhibit pronounced facial corruption and structural inconsistencies, demonstrating that ErodedFace remains effective across different NeRF- and 3DGS-based TFG architectures.}
\label{fig:Qualitative cross-model defense results.}
\end{figure}

As shown in Table~\ref{tab:additional_method_performance}, ErodedFace degrades
reconstruction fidelity, facial structure, and audio-visual
synchronization across all evaluated target models.
The qualitative results in Fig.~\ref{fig:Qualitative cross-model defense results.} further
show that the precise artifact patterns vary across architectures, while
the failure to recover a stable target-specific representation remains
consistent.

The protection therefore transfers across heterogeneous NeRF-based and
3DGS-based pipelines.
This result supports the central design rationale of ErodedFace.
By disrupting semantic information before person specific optimization,
ErodedFace does not rely on a renderer specific weakness and can affect
different personalized representations through their shared upstream
preprocessing dependency.

\begin{figure*}[!tbp]
\centering
\includegraphics[width=1.0\textwidth]{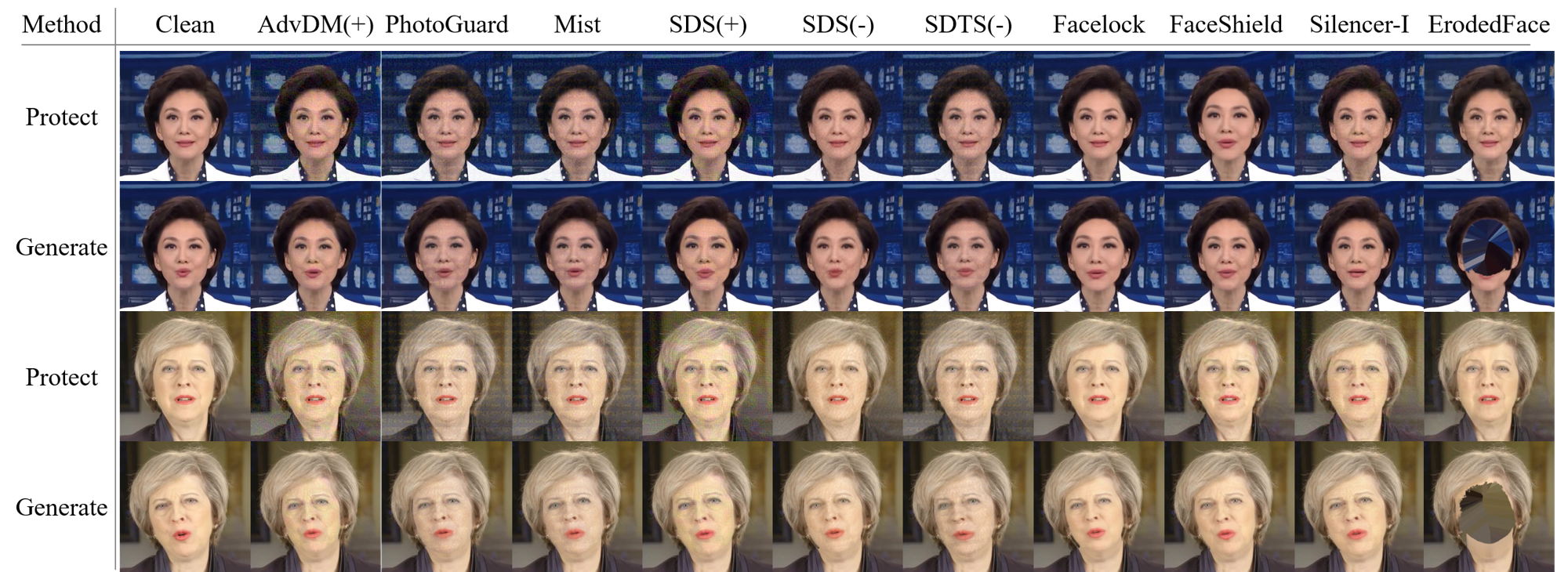}
\caption{\textbf{Qualitative comparison with additional image-level defense methods on SyncTalk.}
For each method, we show the protected reference frames and the corresponding SyncTalk-generated frames. Existing image-level defenses still permit recognizable and structurally plausible synthesis, whereas ErodedFace induces pronounced facial corruption and structural distortions.
}
\label{fig:synctalk_qualitative}
\end{figure*}

\section{Additional Comparisons with Image-Level Defense Methods}
\label{sec:more defense results}

To broaden the comparison beyond the baselines considered in the main experiments, we further evaluate AdvDM(+), Mist, FaceLock, and FaceShield under the same SyncTalk generation pipeline. Each method is applied to the same reference videos, and the resulting protected videos are used for person-specific SyncTalk training. As shown in Table~\ref{more_protect_generted_result}, existing image-level defenses generally produce outputs that remain close to those generated from clean references, as indicated by their relatively high SSIM and PSNR and low FID and M-LMD. In contrast, ErodedFace causes substantially stronger degradation in facial appearance, structural fidelity, and lip motion. Even when some image-level baselines reduce Sync-C, their high SSIM and
PSNR together with low FID and M-LMD indicate that the target appearance
and facial structure remain largely recoverable. In contrast, ErodedFace
simultaneously disrupts visual fidelity, facial structure, and
audio-visual synchronization. The qualitative results in Fig.~\ref{fig:synctalk_qualitative} further confirm this trend that the image-level baselines still yield recognizable and structurally plausible faces, while ErodedFace produces pronounced facial corruption and structural distortions. These results further support our paradigm-mismatch analysis, showing that defenses designed for direct image-conditioned generation do not reliably protect against unauthorized video-based 3D TFG.

\begin{table}[!tbp]
\centering
\small
\setlength{\tabcolsep}{4pt}
\resizebox{\linewidth}{!}{
\begin{tabular}{r | c c c c c}
\toprule
\textbf{Method} & \textbf{SSIM$\downarrow$} & \textbf{PSNR$\downarrow$} & \textbf{FID$\uparrow$} & \textbf{Sync-C$\downarrow$} & \textbf{M-LMD$\uparrow$} \\
\midrule
{AdvDM(+)\cite{xue2023AdvDM+_SDS}} & 0.8820 & 28.53 & 57.84 & 6.7685 & 9.552\\
{Mist\cite{liang2023Mist}} & 0.8949 & 28.53 & 40.41 & 6.0717 & 13.91\\
{FaceLock\cite{wang2025facelock}} & 0.9048 & 29.85 & 16.14 & 7.9346 & 9.897\\
{FaceShield\cite{jeong2025faceshield}} & 0.9130 & 29.68 & 10.37 & 7.8533 & 8.984\\
\midrule
ErodedFace(ours) & \textbf{0.8052} & \textbf{16.27} & \textbf{203.9} & \textbf{0.4276} & \textbf{176.8} \\
\midrule
Ground Truth & 1.00 & $\infty$ & 0.00 & 8.1076 & 0.0000\\
\bottomrule
\end{tabular}}
\caption{\textbf{Additional comparison with image-level defense methods on SyncTalk-generated videos.}
Each method is applied to the same reference videos, and the resulting protected videos are used for person-specific SyncTalk training under identical settings. Lower SSIM, PSNR, and Sync-C and higher FID and M-LMD indicate stronger protection.
}
\label{more_protect_generted_result}
\end{table}

\section{Robustness to Diffusion-Based Purification}
\label{subsec:more anti-purification}

We further evaluate whether diffusion-based purification can remove the protective effect of ErodedFace. To quantify identity preservation after purification, we report ID-S, which measures the average face identity similarity between paired images. Specifically, for each pair of images, we use DeepFace with RetinaFace detection to compare the detected faces under the cosine distance metric. The returned cosine distance is converted into a similarity score, and the final ID-S is obtained by averaging the similarity scores over all valid image pairs where faces are successfully detected. A higher ID-S indicates stronger identity consistency between the generated result and the reference image, while a lower ID-S suggests stronger disruption of identity reconstruction.

\begin{figure}[!t]
\centering
\includegraphics[width=\linewidth]{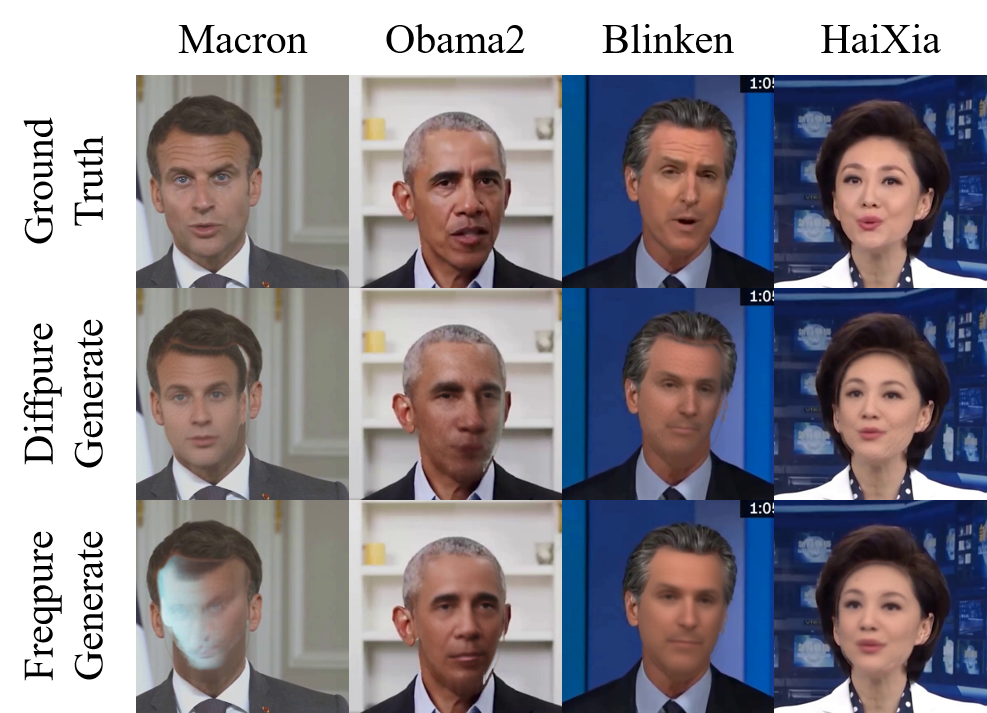}
\caption{\textbf{Qualitative results under diffusion-based purification.}
From top to bottom, we show the purified ErodedFace-protected frame and the corresponding SyncTalk results obtained after DiffPure and FreqPure purification.
}
\label{Qualitative_Comparison_diff_freq_gen}
\end{figure}

Protected videos are first processed by DiffPure~\cite{nie2022diffpure} or FreqPure~\cite{pei2025freqpure} and then used for person-specific SyncTalk training. As shown in Table~\ref{purification_analysis}, purification weakens the protection to some extent, and ErodedFace does not consistently outperform all image-level baselines on every aggregate metric. Nevertheless, its performance remains broadly comparable to the strongest baselines under FreqPure, indicating that the perturbations are not completely eliminated. More importantly, the qualitative results in Figs.~\ref{Qualitative_Comparison_diff_freq_gen} show that noticeable facial artifacts and structural inconsistencies persist after both purification methods. These localized failures are not always fully reflected by full-frame similarity metrics, but substantially reduce the fidelity of the synthesized faces. Overall, the results indicate that ErodedFace retains a meaningful, although partially attenuated, protective effect under diffusion-based purification.

\begin{table}[!htbp]
\centering
\resizebox{\linewidth}{!}{
\begin{tabular}{l l | c c c c}
\toprule
\multicolumn{2}{c|}{\multirow{2}{*}{Method}}
& Diffpure
& Freqpure\\
\multicolumn{2}{r|}{}
& PSNR$\downarrow$/FID$\uparrow$/ID-S$\downarrow$/M-LMD$\uparrow$
& PSNR$\downarrow$/FID$\uparrow$/ID-S$\downarrow$/M-LMD$\uparrow$\\
\midrule
\multicolumn{2}{l|}{AdvDM(-)\cite{liang2023AdvDM-}}
& 29.87/17.19/0.6382/10.78
& 28.94/\textbf{22.51}/\textbf{0.5499}/12.71 \\

\multicolumn{2}{l|}{AdvDM(+)\cite{xue2023AdvDM+_SDS}}
& 29.34/16.79/0.6700/11.79
& 28.75/20.96/0.6043/10.69 \\

\multicolumn{2}{l|}{PhotoGuard\cite{salman2023PhotoGuard}}
& \underline{28.59}/16.50/0.6605/9.629
& \textbf{27.97}/19.32/0.5708/12.09\\

\multicolumn{2}{l|}{Mist\cite{liang2023Mist}}
& 28.73/16.45/0.6404/11.86
& 28.20/19.61/0.5623/\textbf{13.49}\\

\multicolumn{2}{l|}{SDS(+)\cite{xue2023AdvDM+_SDS}}
& 29.00/\underline{17.85}/0.6494/\textbf{12.47}
& 29.76/20.34/0.5886/9.551\\

\multicolumn{2}{l|}{SDS(-)\cite{xue2023AdvDM+_SDS}}
& 29.65/17.80/\underline{0.6258}/\underline{12.42}
& 29.40/\underline{22.48}/\textbf{0.5436}/11.93\\

\multicolumn{2}{l|}{SDST(-)\cite{xue2023AdvDM+_SDS}}
& \textbf{28.55}/\textbf{17.86}/\textbf{0.6257}/11.44
& \underline{28.16}/20.58/0.5511/\underline{12.79}\\

\multicolumn{2}{l|}{Silencer-I\cite{gan2025silence}}
& 29.04/16.61/0.6459/11.99
& 29.49/21.15/0.5673/9.506\\

\midrule
\multicolumn{2}{l|}{ErodedFace(ours)}
& 29.39/15.22/0.6858/8.136
& 29.13/21.68/0.5607/9.421\\
\bottomrule
\end{tabular}}
\caption{\textbf{Robustness to diffusion-based purification.}
Videos protected by each method are processed using DiffPure or FreqPure and subsequently used for person-specific SyncTalk training. Metrics are evaluated on the resulting generated videos. Lower PSNR and ID-S and higher FID and M-LMD indicate stronger residual protection after purification.
}
\label{purification_analysis}
\end{table}

\section{Effect of the Perturbation Budget on Other Preprocessing Modules}
\label{subsec:perturbation_scale}

\begin{figure}[!hbp]
\centering
\includegraphics[width=\linewidth]{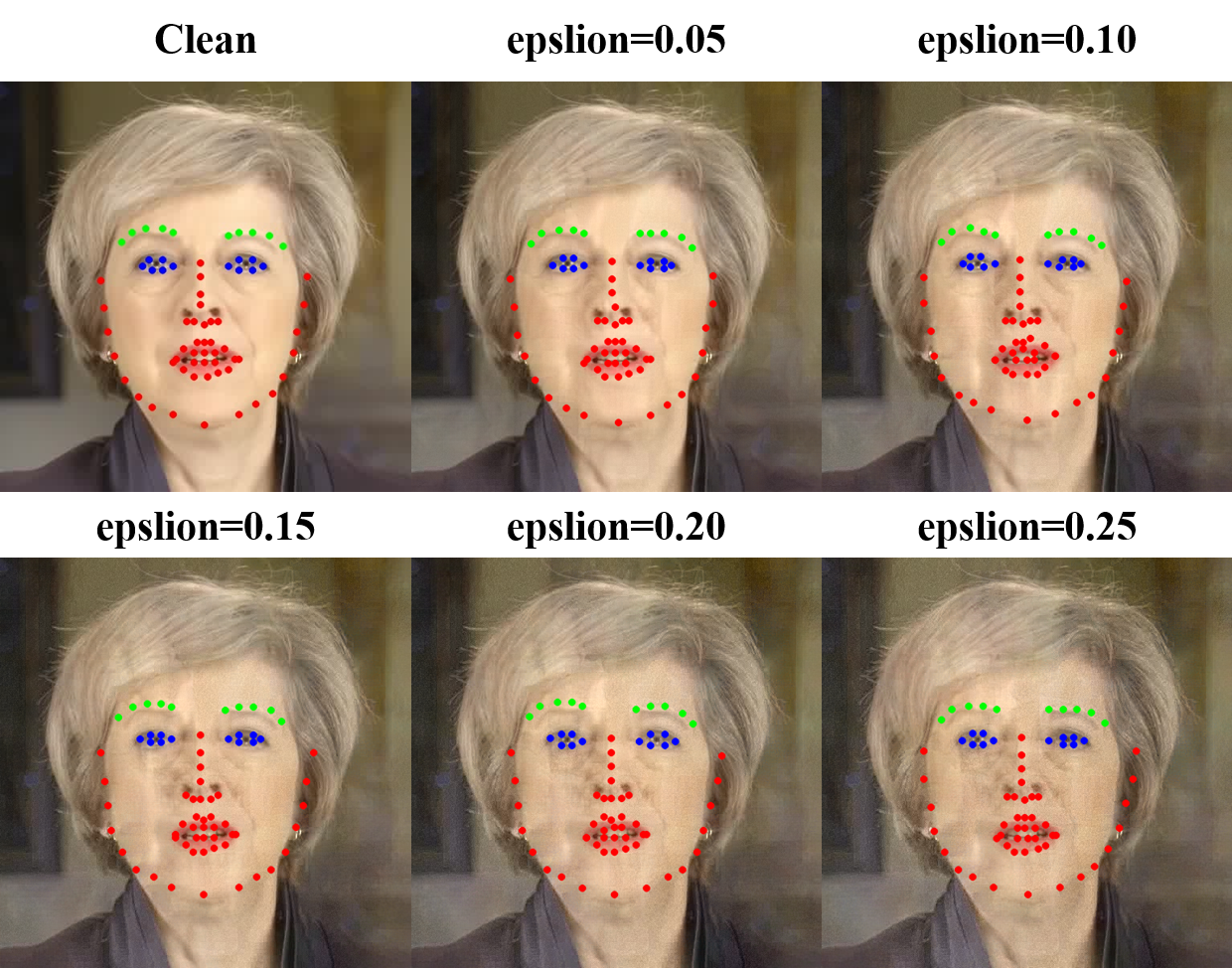}
\caption{
\textbf{Effect of the perturbation budget on facial landmark extraction.}
Although the landmark module is not explicitly targeted during optimization, its predictions become increasingly unstable as the perturbation budget grows. Clear landmark deviations emerge at $\epsilon=0.15$ and become more pronounced under larger budgets.
}
\label{fig:scale_perturbation}
\end{figure}

We further analyze how the perturbation budget affects other
preprocessing modules in 3D TFG pipelines. Specifically, we vary the
maximum perturbation bound $\epsilon$ under the constraint
\[
\|\widetilde I_t-I_t\|_\infty \le \epsilon,
\qquad
\epsilon \in \{0, 0.05, 0.10, 0.15, 0.20, 0.25\}.
\]
As shown in Fig.~\ref{fig:scale_perturbation}, the protected frame remains visually acceptable under small perturbation budgets, while the facial landmark results gradually become unstable as $\epsilon$ increases.

Notably, when $\epsilon$ reaches $0.15$, the 3DMM-related facial processing module already shows clear landmark deviations, even though our optimization objective does not explicitly target this module.
This indicates that disrupting the reference video parsing stage can also affect other personalized information extraction, such as facial geometry estimation and landmark-based alignment.
When the perturbation budget further increases, the facial processing module degrades more severely.
These results suggest that ErodedFace can influence multiple video processing components through a shared perturbation, while a moderate budget is sufficient to balance protection effectiveness and visual quality.

\section{Generalization to Additional Videos}
\label{subsec:additional_data}

To further examine whether ErodedFace generalizes beyond the main evaluation set, we conduct additional experiments on 40 portrait videos from the HDTF dataset~\cite{zhang2021hdtf}. For each video, we compare SyncTalk results generated from the clean reference video with those generated from the ErodedFace protected reference video under the same setting.

As shown in Table~\ref{tab:additional_data_results}, ErodedFace substantially degrades the generated results on the additional HDTF videos. Compared with the clean-reference setting, the protected-reference setting leads to a clear drop in SSIM, PSNR, and Sync-C, together with a large increase in FID. These results indicate that ErodedFace prevents SyncTalk from faithfully reconstructing the target appearance and maintaining audio-visual consistency on previously unseen portrait videos. Therefore, the protective effect is not limited to the main 11 video benchmark, but generalizes to a larger HDTF subset.

\begin{table}[!htbp]
\centering
\small
\setlength{\tabcolsep}{4pt}
\resizebox{\linewidth}{!}{
\begin{tabular}{r | c c c c}
\toprule
\textbf{Reference} & \textbf{SSIM$\downarrow$} & \textbf{PSNR$\downarrow$} & \textbf{FID$\uparrow$} & \textbf{Sync-C$\downarrow$} \\
\midrule
Clean & 0.9320 & 29.65 & 9.872 & 8.09 \\
Protected & 0.6733 & 15.58 & 172.0 & 0.367\\
\bottomrule
\end{tabular}}
\caption{\textbf{Generalization to additional portrait videos.}
We evaluate SyncTalk results on 40 additional portrait videos from HDTF. The clean setting uses the original reference videos, while the protected setting uses ErodedFace protected reference videos. The clear degradation from clean video to protected video indicates that ErodedFace remains effective on the additional HDTF videos.
}
\label{tab:additional_data_results}
\end{table}

\section{Robustness of Frequency Domain Optimization to H.264 Compression.}
\label{app:h264_robustness}

\begin{figure}[h]
\centering
\includegraphics[width=\linewidth]{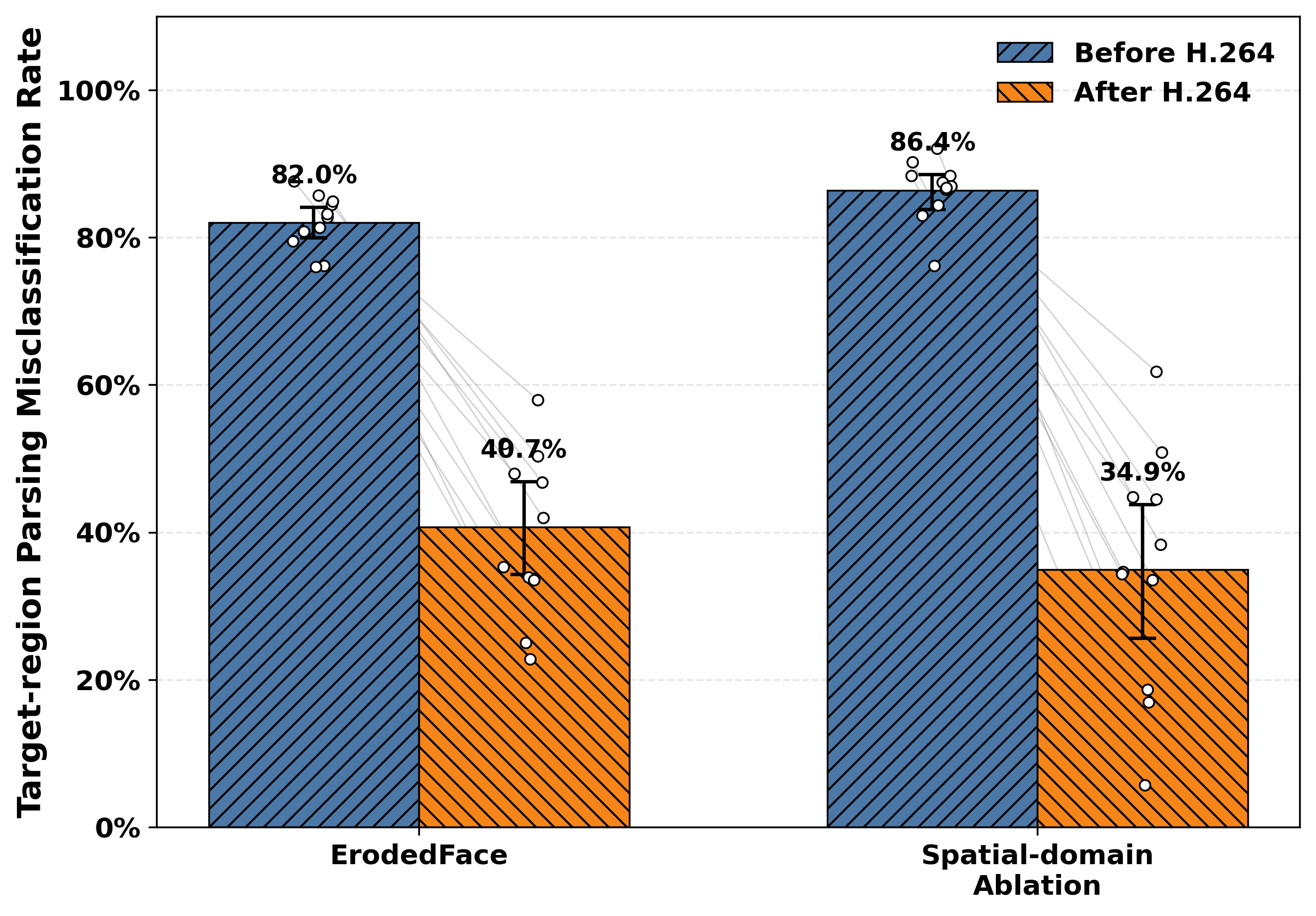}
\caption{Robustness of frequency-domain optimization to H.264
compression.
Bars report macro-averaged rates over 11 identities, points denote
identity-level temporal means, and error bars indicate actual confidence level bootstrap confidence intervals
}
\label{fig:h264_robustness}
\end{figure}

We further evaluate whether the parsing disruption established by
frequency-domain optimization survives practical video recompression.
The spatial-domain setting is an ablation of ErodedFace in which the
frequency-domain perturbation parameterization is replaced by direct
spatial-domain optimization, while all remaining components,
hyperparameters, and optimization budgets follow the same default
configuration.
Consistent with the main experimental setup, both methods use
$\tau=0.8$ and terminate Stage I only after the target-region parsing
misclassification rate reaches at least $80\%$ at all five scales.
Since $\tau$ defines a lower bound rather than an equality target, the
rate at an individual scale can exceed $80\%$.
The native-resolution temporal means shown in
Fig.~\ref{fig:h264_robustness} are therefore expected to be slightly
above the stopping threshold before compression.

The protected videos are re-encoded using H.264 with
\texttt{libx264}, \texttt{preset=medium}, and CRF 18.
H.264 compression is not included in either optimization.
For each identity, we compute the temporal mean target-region parsing
misclassification rate at the native resolution, using the facial
support predicted from the clean video as the fixed evaluation region.
The identity-level results are macro-averaged over all 11 videos, and
uncertainty is estimated using paired identity-level bootstrap.

As shown in Fig.~\ref{fig:h264_robustness}, H.264 compression reduces the
misclassification rate for both parameterizations. ErodedFace starts from
a slightly lower pre-compression rate but retains a higher mean rate after
compression and exhibits a smaller mean paired decrease. These results
indicate that, under the tested setting, frequency-domain
optimization retains more of the targeted parsing disruption than direct
spatial-domain optimization.

\end{document}